\documentclass[letterpaper]{article} % DO NOT CHANGE THIS
\usepackage[submission]{aaai23}  % DO NOT CHANGE THIS
\usepackage{times}  % DO NOT CHANGE THIS
\usepackage{helvet}  % DO NOT CHANGE THIS
\usepackage{courier}  % DO NOT CHANGE THIS
\usepackage[hyphens]{url}  % DO NOT CHANGE THIS
\usepackage{graphicx} % DO NOT CHANGE THIS
\usepackage{natbib}  % DO NOT CHANGE THIS AND DO NOT ADD ANY OPTIONS TO IT
\usepackage{caption} % DO NOT CHANGE THIS AND DO NOT ADD ANY OPTIONS TO IT
\usepackage{algorithm}
\usepackage{algorithmic}
\usepackage{mathrsfs}
\usepackage{amssymb}
\usepackage{bm}
\usepackage{amsmath}
\usepackage{dsfont}
\usepackage{booktabs}
\usepackage{epstopdf}
\usepackage{multirow}
\usepackage{comment}
\usepackage[table,xcdraw]{xcolor}
\usepackage{cite}
\usepackage{endnotes}
\usepackage{makecell}
\usepackage{epstopdf}
\usepackage{newfloat}
\usepackage{listings}
\usepackage{endnotes}
\usepackage{makecell}

\DeclareCaptionStyle{ruled}{labelfont=normalfont,labelsep=colon,strut=off} % DO NOT CHANGE THIS
\floatstyle{ruled}
\newfloat{listing}{tb}{lst}{}
\floatname{listing}{Listing}

\title{GANN: Graph Alignment Neural Network for Semi-Supervised Learning}
\author{Linxuan Song,\textsuperscript{\rm 1} Wenxuan Tu,\textsuperscript{\rm 1}, Sihang Zhou,\textsuperscript{\rm 2} Xinwang Liu,\textsuperscript{\rm 1} \footnotemark[2] En Zhu,\textsuperscript{\rm 1} \footnotemark[2]}
\affiliations{
    \textsuperscript{\rm 1}College of Computer, National University of Defense Technology \\
    \textsuperscript{\rm 2}College of Intelligence Science and Technology, National University of Defense Technology\\
}

\usepackage{bibentry}
\begin{document}
\maketitle

\begin{abstract}

\footnotetext[2]{Corresponding author}

Graph neural networks (GNNs) have been widely investigated in the field of semi-supervised graph machine learning. Most methods fail to exploit adequate graph information when labeled data is limited, leading to the problem of oversmoothing. To overcome this issue, we propose the Graph Alignment Neural Network (GANN), a simple and effective graph neural architecture. A unique learning algorithm with three alignment rules is proposed to thoroughly explore hidden information for insufficient labels. Firstly, to better investigate attribute specifics, we suggest the feature alignment rule to align the inner product of both the attribute and embedding matrices. Secondly, to properly utilize the higher-order neighbor information, we propose the cluster center alignment rule, which involves aligning the inner product of the cluster center matrix with the unit matrix. Finally, to get reliable prediction results with few labels, we establish the minimum entropy alignment rule by lining up the prediction probability matrix with its sharpened result. Extensive studies on graph benchmark datasets demonstrate that GANN can achieve considerable benefits in semi-supervised node classification and outperform state-of-the-art competitors.

\end{abstract}

\section{1 \hspace{0.3cm}Introduction}
Graph semi-supervised learning is an important research topic in the era when graph data rapidly accumulate while data labeling is unaffordably expensive and time-consuming. To provide good performance with little human guidance, graph semi-supervised algorithms make large efforts to exploit hidden information within the data. However, lacking of enough supervision further aggravates the oversmoothing problem of graph neural networks. As an example, graph convolutional networks (GCNs)\cite{1} use the laplacian matrix to extract features and excel at link prediction and clustering. When employing more than two layers of GCN, it may fuse node features from different clusters and make them hard to tell apart, i.e., oversmoothing.

The existing methods usually exploits either structural information or attribute information to eliminate the oversmoothing problem.
Inspired by the success of image restoration, Graph Completion\cite{21} conceals specific nodes of the input graph by deleting their features. Additionally, AttributeMask\cite{f_2,f_3} aims to reconstruct the dense feature matrix processed by principal component analysis. By mining the attributes, each of the above methods can produce a less collapse-prone node representation. The following papers pertain to the extraction of structural information. APPNP\cite{7} uses personalized pagerank with GCNs and comes up with a sophisticated global propagation scheme. Based on APPNP, the ADAGCN\cite{11} incorporates AdaBoost into graph neural networks for multilayer aggregation, which can employ higher-order neighbor knowledge for node representation construction.

Although the oversmoothing problem has been largely eliminated by the mentioned methods, they fail to exploit the two kinds of essential information for further performance improvement. Firstly, the majority of GNN models do not associate the representation with the original attributes following multiple convolutions. They pay less attention to the relationship between the final node's representation and the original data, resulting in the underutilization of characteristics. Secondly, the vast majority of the aforementioned articles can only aggregate first- or second-order neighbor information using the adjacency matrix. Nonetheless, a number of investigations have demonstrated that the adjacency matrix's node connections are not always meaningful, and nodes from different clusters may be connected. This necessitates models with the capacity to dig deep into the structural information in order to build more trustworthy node representations.

To solve the mentioned problems, we propose a generalised graph alignment neural network (GANN) on the basis of the ADAGCN\cite{11} learning framework. It aligns the intermediate network representations with three kinds of information landmarks for hidden supervision exploitation.

To fully investigate the attributes, we first propose the feature alignment rule based on the notion of graph autoencoder. We enrich node representation by aligning the feature correlation matrix with the embedding correlation matrix. Then, to maximize the utilization of the multi-hop graph structure and labeling information, we suggest the cluster center alignment rule. We compute the inner product of cluster-centered embedding and align it with the unit matrix. By minimizing intra-cluster distance and eliminating inter-cluster noise, we prevent the model from becoming too smooth during deep training. Finally, we present the minimum entropy alignment rule to generate more reliable outcomes in semi-supervised tasks. We force the model to generate predictions with low entropy by aligning the prediction probability matrix with its sharpened results\cite{45}. In the experimental section, we demonstrate the effectiveness of GANN through comprehensive algorithm comparisons and ablation studies. Experimental results show that our proposed model, GANN, outperforms the state-of-the-art approach on the benchmark datasets.

The contributions of this paper are as follows:

\begin{itemize}
\item We propose GANN, a model that can alleviate the oversmoothing phenomenon in graph neural networks by fully exploiting the essential information of graphs.

\item We propose three alignment rules: feature alignment rule, cluster center alignment rule, and minimum entropy alignment rule. From the views of the attribute, structural, and prediction result optimization, they can leverage graph information and labeled data to generate quality node representations.
%They can utilize graph knowledge and labeled data from the perspectives of attribute, structural, and prediction result optimization, respectively.

\item We perform node classification experiments on the benchmark datasets with labels of varying labeling rates, and all datasets offer improved results. In addition, the validity of alignment rules is examined to verify that the proposed GANN can adequately use the essential information of graph data.

\end{itemize}

\section{2\hspace{0.3cm}Related Work}
\subsection{Graph Semi-Supervised Learning}
Semi-supervised tasks\cite{pr_2, pr_3} require only a small quantity of labeling information, making full use of the vast amount of unlabeled information, and can offer substantial improvements in practice. Graph semi-supervised\cite{pr_1} learning has gained a lot of attention due to its unique structure and wide range of applications. There are some semi-supervised algorithms\cite{pr_4, pr_5, pr_6} that can partially address the problem of oversmoothing phenomenon. 

%There are some algorithms\cite{ICRN, GCC_LDA, CCGC, ICL_SSL, liuyue_DCRN, liuyue_HSAN, liuyue_SCGC} that can partially address the problem of oversmoothing phenomenon. 

GCN-Cheby\cite{c-gcn} uses CNNs in spectral graph theory to create fast localized convolutional filters on graphs. GAT\cite{12} uses masked self-attentional layers to improve graph convolution-based algorithms. Jumping knowledge networks with concatenation (JK-Net)\cite{43}: introduces jump connections into the final aggregation mechanism to extract knowledge from different graph convolutional layers. GPRGNN\cite{gprgnn} introduces a new Generalized PageRank GNN architecture that adaptively trains GPR weights to concurrently improve node feature and topological information extraction independent of node label homophily or heterophily. MixupForGraph\cite{mixup} utilizes the representations of each node's neighbors prior to Mixup for graph convolutions. SGC\cite{sgc} simplifies models by reducing nonlinearities and collapsing weight matrices between layers. GWNN\cite{gwnn} uses graph wavelet transform to improve on graph Fourier-based spectral graph CNN algorithms. PPNP and APPNP\cite{7} rely mainly on the PageRank concept to solve the deeper adjacency information, which is difficult to obtain with graph neural networks. Experiments are conducted on several datasets with varying labeling rates to evaluate the efficacy of our proposed model in comparison to the aforementioned related methods.

\subsection{Graph Neural Networks}
GCN\cite{1} is the first method to take CNN from euclidean to graph domains. It can be explained in both spectral and spatial fields, leading to the derivation of articles in two directions. Common spatial fields include GAT\cite{12}, TAAGCN\cite{pr_7}, etc. In the spectral domain, such as FAGCN\cite{8}, NFCGCN\cite{pr_8}, etc. However, none of the methods with GCN as the base component can exploit the higher-order neighbor information because of the collapse phenomenon after their multilayer aggregation. 

Currently, multilayer perceptrons (MLPs) are widely utilized by the public due to their adaptability and efficacy. Zhu\cite{31} develops MLP-based networks which perform comparably to visual networks in the field of images. Experimental validation in Zhang\cite{35} demonstrates that both GNN and MLP have sufficient expressiveness when the dimension of the representation is sufficiently large. We applied a simple MLP layer as the main structure in the model. It is proved that good feature mining results can be achieved with MLP only. 

GAE\cite{16} is one of the prevalent methods for generating tasks. It employs GCN to encode a model implicitly and the decoder uses inner products. To rebuild the original graph structure, the inner product of the implicit representation is aligned with the original adjacency matrix. Graph autoencoders have begun to be widely explored, with several models employing their concepts. VGAE\cite{16} extends GAE's variational autoencoder. Later, SIG-VAE\cite{17} considers hierarchical variational inference to generate graph data. ARGA/ARVGA\cite{18} standardized GAE/VGAE employing GANs. We refer to the idea of adjacency matrix alignment in the traditional structure of GAE for model design and apply it to the part of feature data mining. 

\subsection{Oversmoothing Phenomenon}
Oversmoothing is generated by the convolution of many layers of GCN, with Li\cite{23} first proposing it. The node representation of various clusters is too close, affecting model performance. Xu\cite{24} mentions that the oversmoothing speed among various types of nodes varies. More edge nodes slow down oversmoothing. The solution in Kipf\cite{1} and Xu\cite{46} is based on ResNets\cite{4}, which act by adding residual connections. But the model is less effective. Additionally, there are some algorithms\cite{ICRN, GCC_LDA, CCGC, ICL_SSL, liuyue_DCRN, liuyue_HSAN, liuyue_SCGC, liuyue_survey, SymCL, ABSLearn, AKGR} that can alleviate the problem of oversmoothing phenomenon and introduce some of the applications of GNN. 

By aggregating the first-order neighbors of different frequencies, there is also an innovation in frequency. FAGCN\cite{8} makes the central node more informative and increases the difference from other nodes, i.e., by aggregating both high and low-frequency information. ADAGNN\cite{9} grades the frequencies and adaptively aggregates them. Another kind of feature extraction of graph data does not use the form of convolution. Instead, they simply employ the ordinary nonlinear activation function to process the data, such as the common MLPs and other model structures. Several studies attempt to tackle the challenge posed by Laplace smoothing via transforming the filter form\cite{10}. The structure of our designed model aims to mitigate the impact of the oversmoothing problem in terms of node representation generation. 

\begin{table}[t]
	\centering
        \small
        \scalebox{1.0}{
	\begin{tabular}{l|l}
		\toprule
		\multicolumn{1}{c|}{Notations} & \multicolumn{1}{c}{Meaning} \\ 
		\midrule
            $N$&    Sample Number   \\
            $d$&    Feature Dimension   \\
            $h'$&   Embedding Dimension   \\
            $L$&    Layer Number   \\
            $C$&    Cluster Number   \\
		$\textbf{X}\in \mathbb{R}^{N\times d}$&    Feature Matrix                          \\
		$\textbf{A}\in \mathbb{R}^{N\times N}$&    Adjacency Matrix                          \\
		$\textbf{I}\in \mathbb{R}^{N\times N}$&      Identity Matrix                        \\
		% $\hat{\tilde{\textbf{A}}}\in \mathbb{R}^{n\times n}$&  Normalized Adjacent Matrix                            \\
		$\hat{\textbf{A}}\in \mathbb{R}^{N\times N}$&   Normalized Adjacent Matrix                       \\
		$\textbf{D}\in \mathbb{R}^{N\times N}$&    Degree Matrix                         \\
            $\hat{\textbf{X}}\in \mathbb{R}^{N\times d}$&   Normalized Feature Matrix       \\
            ${\textbf{H}}_1^{(l)}\in \mathbb{R}^{N\times h'}$&   Node Embedding for the Layer $l$       \\
		${\textbf{F}}^{'}\in \mathbb{R}^{N\times N}$& Feature Correlation Matrix       \\
            $\hat{\textbf{F}}\in \mathbb{R}^{N\times N}$& Embedding Correlation Matrix       \\
            $\bar{\textbf{E}}\in \mathbb{R}^{C\times h'}$& Cluster Center Matrix \\
            $\hat{\textbf{E}}\in \mathbb{R}^{h'\times h'}$&  Cluster Center Correlation Matrix \\
            
            $\textbf{Z}\in \mathbb{R}^{N\times C}$&  Prediction Probability Matrix  \\
		\bottomrule
	\end{tabular}}
 \caption{Notation Summary}
	\label{tab:notation}
\end{table}

\section{3 \hspace{0.3cm}Our Proposed Model: GANN}
In the following chapter, we will first discuss the notations used and task definition, then describe our motivation and the details of the GANN components.

\subsection{Preliminaries}
\subsubsection{Symbols Definition}
Given an undirected graph \begin{math}\mathcal{G}=(V,\mathcal{E} ,X)\end{math}, with adjacency matrix $\textbf{A} \in \mathbb{R}^{N\times N}$, where $V$ is the set of nodes which has $N$ samples. $\mathcal{E} $ is the set of edges. $\textbf{X}\in \mathbb R^{N\times d}$ is the attribute matrix. The Laplacian matrix of the graph is defined as $\textbf{L}=\textbf{D}-\textbf{A}$, where $\textbf{D}\in \mathbb R^{N\times N}$ is a diagonal degree matrix with $\textbf{D}_{ii}=\sum\nolimits_j^{}{\textbf{A}_{ij}}$. The symmetric normalized Laplacian matrix are defined as $\textbf{L}_{sym}=\textbf{I}-\textbf{D}^{-\frac{1}{2}}\textbf{AD}^{-\frac{1}{2}}$. 
%$\textbf{L}_{sym}=\textbf{D}^{-\frac{1}{2}}\textbf{LD}^{-\frac{1}{2}}$, i.e.,  
We use $\hat{\textbf{A}}=\tilde{\textbf{D}}^{-\frac{1}{2}}\tilde{\textbf{A}}\tilde{\textbf{D}}^{-\frac{1}{2}}\in \mathbb R^{N\times N}$ as a frequency filter where $\tilde{\textbf{A}}=\textbf{A}+\textbf{I}$. $\tilde{\textbf{D}}$ is the degree matrix of $\tilde{\textbf{A}}$ and $\textbf{I}$ denotes the identity matrix. We first normalize $\textbf{X}$ by calculating $\hat{\textbf{X}}=Normalize(\textbf{X})$ and then take the resultant $\hat{\textbf{X}}$ matrix as input.
%We use $\hat{\textbf{X}}=Normalize(\textbf{X})$ as feature input.

\subsubsection{Task Definition}
Graph alignment neural network for semi-supervised learning. 
%Train a neural network to provide reliable representations of the node with few labeled data. 
The model receives $\textbf{X}$ and $\textbf{A}$ as inputs, performs iterative training of several layers and then averages the output $\textbf{Z}$ of each layer as the final results.

\begin{figure*}[t]
\centering
\includegraphics[scale=0.60]{ 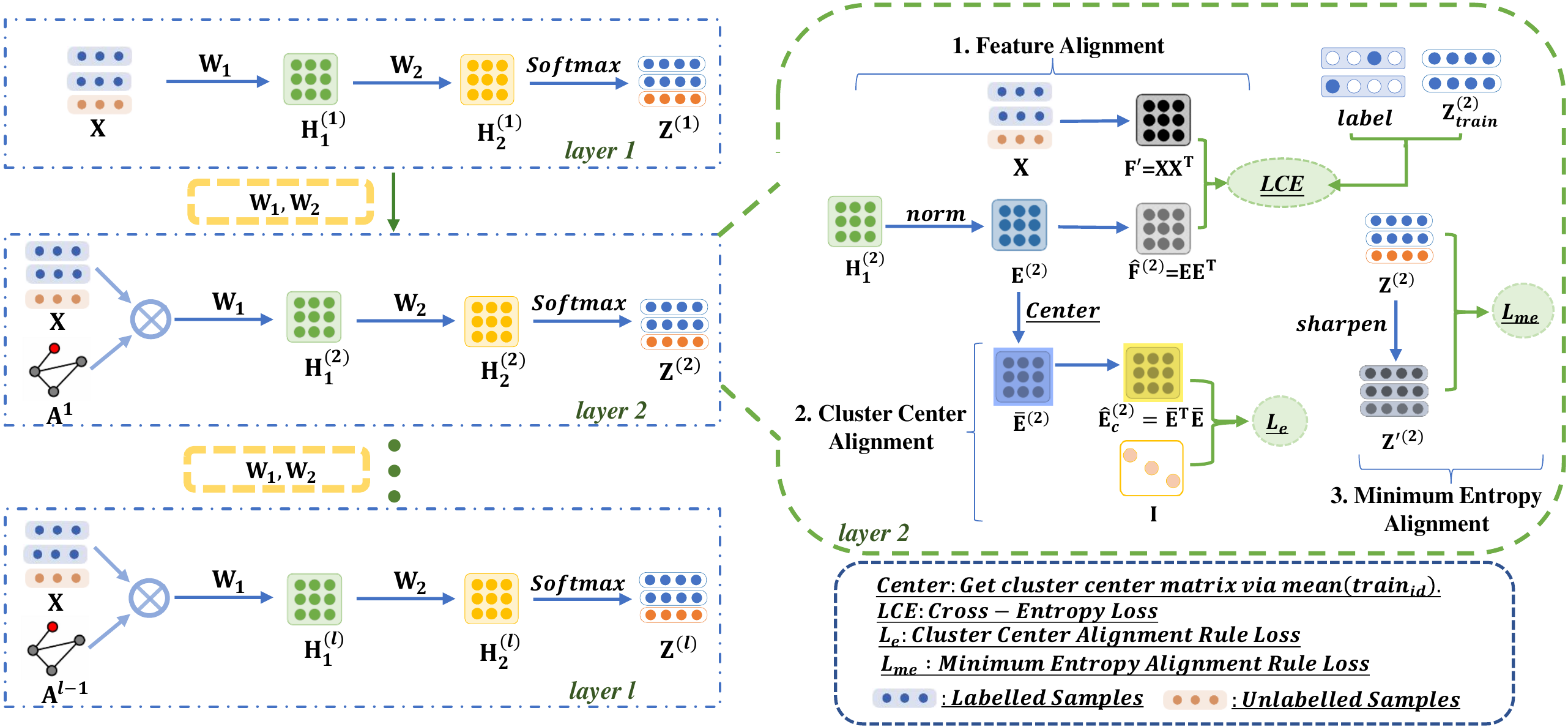}
\caption{Illustration of GANN. The model is trained using sequential iteration, as depicted on the figure left. We utilize one MLP layer for feature extraction and one MLP layer with softmax for probability calculation. The parameters for each model structure layer are shared. In addition, the model includes three alignment rules to optimize the results, figure right. Inputs to the model are the attribute matrix and the adjacency matrix of various hops, outputs are the cluster probabilities.}
\label{figure1}  
\end{figure*}

\subsection{Motivation}
\textbf{ADAGCN.}
An efficient model that combines the data of nodes and their neighbors by integrating the traditional graph neural network with Adaboost. Specifically, the single-layer model structure of ADAGCN\cite{11} consists of MLPs, the training is usually multilayered. The $l$-th layer's input is the $l$-hop adjacency matrix of the graph $\mathcal{G}$ and the attribute matrix. Each layer of the network is structure-identical with shared parameters. In ADAGCN, the result $\textbf{Z}$ is given at the end of each layer. After all training layers have been conducted, $\textbf{Z}$ is then weighted to generate the final results $\textbf{Z}'$. The loss function of ADAGCN is formulated as:

\begin{equation}
\mathcal{L}_{semi}=-\frac{1}{n}\sum_{i}{w_i[ y_i\ln \hat{y}_i+( 1-y_i ) \ln ( 1-\hat{y}_i ) ]},
\end{equation}
where $n$ represents the number of labeled samples, $y_i$ and $\hat{y}_i$ represent the predicted label and the true label, respectively. $w_i$ denotes the learnable weight. ADAGCN has been proven superior to GCN in addressing the issue of embedding collapse.

\textbf{Limitations.}
Despite the effectiveness of ADAGCN, we observe that it has the following two limitations: i) Overlooking the underlying meaningful attributes. On the one hand, ADAGCN takes only the attribute matrix $\textbf{X}$ as input during the first layer, which results in ignoring the attributes of unlabeled nodes according to its objective function. On the other hand, the adjacency matrix is multiplied with the attributes to correlate non-training samples. This way only uses neighborhood-level knowledge from the adjacency matrix, which is limited for constructing node representations. ii) Ignoring the significant higher-order neighbors. More precisely, after multiplication, adjacency matrix becomes denser. This causes more neighbor noise due to erroneous edges, hindering the model from mining higher-order information in depth.
To solve the above issues, we propose GANN to explore the essential information of the graph via three alignment rules, as depicted in Figure \ref{figure1}.

\subsection{Alignment Rules}
In this section, we will introduce three alignment rules of GANN in detail, i.e., feature alignment rule, cluster center alignment rule, and minimum entropy alignment rule.

\subsubsection{Feature Alignment Rule}
As stated previously, existing models cannot fully leverage original graph attributes. To tackle this problem, we introduce the feature alignment rule, where we align the feature correlation matrix with embedding correlation matrix. Specifically, we first calculate the similarity matrix $\textbf{S}$ via $\hat{\textbf{X}}$ to obtain the $0$-$1$ feature correlation matrix $\textbf{F}'$. In the process of similarity estimation, we adopt the jaccard, cosine, and gaussian kernel approaches. Cosine similarity is finally employed according to the performance comparison, formulated as:

\begin{equation}
\centering
\begin{aligned}
    \mathbf{S}_{ij}&=\frac{\hat{\mathbf{x}}_i}{\| \hat{\mathbf{x}}_i \| _2}\cdot \frac{\hat{\mathbf{x}}_j}{\| \hat{\mathbf{x}}_j \| _2},
\end{aligned}
\end{equation}

\begin{equation}
\centering
\begin{aligned}
    \mathbf{f}_{ij}&=\begin{cases}	\mathbf{S}_{ij}, \mathbf{S}_{ij}\geqslant \eta \\
    0,\hspace{0.3cm}\mathbf{S}_{ij} \hspace{0.1cm} \textless \hspace{0.1cm}  \eta \,\,\\\end{cases},
    \end{aligned}
\end{equation}

\begin{equation}
\centering
\begin{aligned}
    \mathbf{f}_i'&=topk(\mathbf{f}_i)\hspace{0.1cm},
\end{aligned}
\end{equation}
where $\cdot$ denotes the dot product and $||$ is the $L_2$ parametrization. $\mathbf{S}_{ij}$ is an element of the similarity matrix $\textbf{S}$, which represents the similarity between node $i$ and node $j$. Since the underlying graph structure is sparse, we set the elements of $\textbf{S}$ that are smaller than a threshold $\eta $ to $0$. In addition, $topk$ indicates that $k$ nodes with the highest similarity value of node $i$ are selected, and the rest are assigned to $0$.

Next, we obtain the embedding correlation matrix $\hat{\textbf{F}}$ by performing the inner product of the embedding matrix $\textbf{E}$. Specifically, $\hat{\textbf{X}}$ is first layer input, while the smoothed adjacency matrix $\hat{\textbf{A}}$ and $\hat{\textbf{X}}$ form the second layer's input. After the input, the dropout layer is used to increase the model's robustness, and then the MLP layer is employed to generate the embedding matrix, shown below.

\begin{equation}
\textbf{H}_{1}^{{(l)}}=\mathrm{\sigma_1}( \mathrm{Dropout}( \hat{\textbf{X}}\hat{\textbf{A}}^{( l-1 )} ) \textbf{W}_1 ), 
\label{e-2}
\end{equation}

\begin{equation}
\mathbf{e}_{i}^{(l)}=\mathrm{ }\frac{\mathrm{h}_{1i}^{(l)}}{\max \!\:( \| \mathrm{h}_{1i}^{(l)} \| _2,\mathrm{  }\epsilon )}\mathrm{ },
\end{equation}

\begin{equation}
    \hat{\textbf{F}}^{(l)}=\sigma_2( \textbf{E}^{(l)}{\textbf{E}^{(l)}}^{\top}),
    \label{e6}
\end{equation}
where $l$ denotes the model's layer number. $l=1,...,L$. $\textbf{W}_{1}\in \mathbb{R}^{N\times d}$ is a trainable parameter and $\textbf{H}_{1}^{(l)}\in \mathbb{R}^{N\times h'}$ is the hidden node representation for the layer $l$. $\textbf{H}_{1}^{(l)}$ contains $h'$ cells and $\mathbf{h}_1^{(l)}$ is normalized to generate $\mathbf{e}_{i}^{(l)}$, the final node embeddings. $\sigma_1$ and $\sigma_2$ denote the relu and sigmoid activation functions, respectively. $\epsilon $ is the minimal value.

Lastly, we calculate Kullback-Leibler divergence between the feature correlation matrix $\textbf{F}'$ and embedding correlation matrix $\hat{\textbf{F}}$ to measure their difference. As $\textbf{F}'$ is fixed, so is its entropy value. Consequently, it can be written as a cross-entropy loss function, as follows,

\begin{equation}
    {\mathcal{L}_f}^{(l)}=CE(\hat{\textbf{F}}, \textbf{F}'),
\label{e7}
\end{equation}
where $CE(\cdot)$ denotes the Cross-Entropy loss\cite{celoss}.
By applying this rule and minimizing Eq.\eqref{e7}, the necessary node characteristic for subsequent tasks is fused into the node representation in order to enrich it.

\subsubsection{Cluster Center Alignment Rule}
Although the feature alignment rule could efficiently employ attribute information to generate node representations, it suffers from the oversmoothing issue. Therefore, we develop the cluster centre alignment rule to fully utilize structural information. In detail, we utilize a few labeled nodes to calculate their cluster-centered embedding matrix $\bar{\textbf{E}}$ and align its inner product with a unit matrix via Kullback-Leibler divergence. For instance, we assume that Cora-ML has seven clusters and each cluster contains $20$ labeled samples, the average embedding of $20$ samples denotes the cluster-center embedding, which is formulated as follows:

%(尽管特征对齐XXX，然而它有哪些欠缺，所以引出cluster center 对齐,A和B怎么获得的，A和B对齐，对齐的方式又是什么,为什么做这个对齐，有什么潜在的好处). 
 
%To measure the current cluster-centered representation's quality.
\begin{equation}
    {\bar{\mathbf{e}}}_{i}^{(l)}=\frac{\sum\nolimits_{j\in i}^m{{\mathbf{e}_j}^{(l)}}}{m},
\label{e9}
\end{equation}

\begin{equation}
    {\hat{\textbf{E}}}^{(l)}=\bar{\textbf{E}}^{{(l)}^{\top}}{\bar{\textbf{E}}}^{(l)},
\label{e10}
\end{equation}
where $i$ denotes the current cluster and $m$ means the number of samples in the cluster $i$. ${\bar{\textbf{E}}}^{(l)}\in \mathbb{R}^{C\times h'}$ and ${\hat{\textbf{E}}}^{(l)}\in \mathbb{R}^{h'\times h'}$ is the inner product of $\bar{\textbf{E}}^{(l)}$. $C$ is the number of clusters. 

We align the unit matrix with ${\hat{\textbf{E}}}$ since the cluster center should keep closer to itself and far away from the other clusters. The loss function is formulated as:

\begin{equation}
    \mathcal{L}_e=\frac{\sum\nolimits_i^C{( 1-\frac{{\hat{\mathbf{E}}}^{(l)}_{ii}}{C} ) ^2}}{C}+\lambda \frac{\sum\nolimits_i^C{\sum\nolimits_{j\ne i}^C{( \frac{{\hat{\mathbf{E}}}^{(l)}_{ij}}{C}^{} ) ^2}}}{C(C-1)}.
    \label{e11}
\end{equation}

In Eq.\eqref{e11}, $\lambda$ is a hyper-parameter that is employed to coordinate the cluster-center embeddings with other noise, which alleviates the oversmoothing issue by using cluster centers with labeled data. In our view, deterministic criteria improve model accuracy. Hence, it gets rid of the problem that the node representation cannot make full use of the graph structure information.

\subsection{Minimum Entropy Alignment Rule}
The first two alignment rules optimize node representations by mining graph attribute and structure data. However, it still poses a great challenge to create an accurate prediction probability matrix with a small number of labels for semi-supervised tasks. To overcome this problem, we propose a minimum entropy alignment rule by aligning the prediction probability matrix and its sharpened\cite{45} results. Specifically, we first calculate $\textbf{H}_2^{(l)}$ using an MLP layer to ensure that the dimensions of the learned embeddings and clusters are constant. The prediction probability matrix $\textbf{Z}^{(l)}\in \mathbb{R}^{N\times C}$ is then generated using softmax. Moreover, we calculate the cluster relationship probabilities of all samples through a sharpening operation to get the sharpened matrix $\hat{\mathbf{Z}}$.
\begin{equation}
    \textbf{H}_{2}^{(l)}=\sigma_1 ( \textbf{H}_{1}^{(l)}\textbf{W}_2 ),
    \label{e12}
\end{equation}

\begin{equation}
    \mathbf{z}_{i}^{(l)}=\frac{\mathbf{e}_{}^{h_{2i}^{(l)}}}{\sum_{j=1}^N{e^{h_{2j}^{(l)}}}},
\end{equation}

\begin{equation}
    \hat{\mathbf{Z}}_{ij}^{(l)}=\exp ( {\mathbf{Z}_{ij}^{(l)}}^{\frac{1}{tem}} ) /\sum_{c=1}^C{\exp ( {\mathbf{Z}_{ic}^{(l)}}^{\frac{1}{tem}} )}.
    \label{sharpen}
\end{equation}

In Eq.\eqref{sharpen}, the cluster of node $i$ indicates $c$, ranging from $1$ to $C$. $tem$ refers to a hyper-parameter, which controls the sharpness of the categorical distribution. Next, we decrease the distance between the prediction probability matrix $\textbf{Z}$ and the sharpened result $\hat{\mathbf{Z}}$ to optimize the network. The rule loss function is formulated as:

\begin{equation}
    \mathcal{L}_{me}=\frac{1}{N}\sum_{i=1}^N{||\mathbf{z}_{i}^{(l)}-\hat{\mathbf{z}}_{i}^{(l)}||_{2}^{2}}.
\end{equation}

Grandvalet\cite{44} has shown that unlabeled examples are mostly beneficial when clusters have small overlap. We optimize the results by reducing the entropy of the prediction, which increases the confidence of the classification results.

% % ALGORITHM
\begin{algorithm}[t]
\small
\caption{GANN}
\label{ALGORITHM}
\flushleft{\textbf{Input}: Normalized adjacent matrix $\hat{\textbf{A}}$, normlized attribute matrix $\hat{\textbf{X}}$. Number of layers $L$, model $\mathbf{f}_{mlp}(\textbf{X}, \Theta)$ and patience step $p$.} \\
\flushleft{$\textbf{Output}$: Prediction $\textbf{Z}$.}
\begin{algorithmic}[1]
\STATE Initialization: $\textbf{X}^{(0)}=\hat{\textbf{X}}$, compute the feature correlation matrix $\textbf{F}=\hat{\textbf{X}}\hat{\textbf{X}}^{\top}$, iteration step $s=0$.
\FOR{$l$ from $0$ to $L$}
\WHILE{ $s<p$ }
\STATE Obtain node embedding: $\textbf{H}^{(l)}=\mathbf{f}_{mlp}(\textbf{X}^{(l)}, \Theta_1)$.
\STATE Get the embedding correlation matrix via Eq.\eqref{e6}: $\hat{\textbf{F}}^{(l)}=\textbf{H}^{(l)}{\textbf{H}^{(l)}}^{\top}$.
\STATE Compute the cluster centers of training nodes in Eq.\eqref{e9}, $\bar{\textbf{E}}^{(l)}=center(\textbf{H}^{(l)}, train_{id})$, and calculate its corresponding inner product $\hat{\textbf{E}}^{(l)}$ in Eq.\eqref{e10}.
\STATE Calculate probability prediction matrix: $\textbf{Z}^{(l)}=Softmax(\mathbf{f}_{mlp}(\textbf{H}^{(l)}, \Theta_2))$.
\STATE Update model by minimizing the loss $\mathcal{L}$ in Eq.\eqref{e18}.
\IF{$\mathcal{L}_{s} \geq \mathcal{L}_{s-1}$}
\STATE $s+=1$
\ENDIF
\ENDWHILE
\ENDFOR
\STATE \textbf{return} $\textbf{Z}=\frac{\textbf{Z}^{(1)}+\cdots +\textbf{Z}^{(l)}+\cdots+\textbf{Z}^{(L)}}{L}$ by Eq.\eqref{e17}.
\end{algorithmic}
\end{algorithm}

\subsection{Overview of Our Proposed GANN}
In this subsection, we formally deﬁne the whole architecture of GANN. The mathematical formulation of GANN is defined as:

\begin{equation}
\centering
\begin{aligned}
\begin{cases}
&\textbf{H}_{1}^{(l)}=f_1( \hat{\textbf{X}}\hat{\textbf{A}}^{( l-1 )} ),\\
&\textbf{H}_{2}^{(l)}=f_2( \textbf{H}_{1}^{(l)} ),\\
&\textbf{Z}^{(l)}=\log ( Softmax ( \textbf{H}_{2}^{(l)} ) ),\\
% \end{split}
% \label{e16}
% \end{equation}
% \begin{equation}
&\textbf{Z}=\frac{\textbf{Z}^{( 1 )}+ \cdots +\textbf{Z}^{(l)}+ \cdots +\textbf{Z}^{(L)}}{L},
\end{cases}
\end{aligned}
\label{e17}
\end{equation}
where $f_1$ denotes the nonlinear function part of the above Eq.\eqref{e-2}, $f_2$ denotes the function part in Eq.\eqref{e12}. $l$ is the number of the current network layer, ranging from $1$ to $L$. The parameters of each model layer are shared and updated sequentially. Note that we adopt the optimal learned weight matrix of the $l$-th layer to initialize that of the $(l+1)$-th layer. The training optimal weight coefficients of the previous layer initialize the next. The final loss function is formulated as: 

\begin{equation}
    \mathcal{L} =\mathcal{L}_{semi} +\mathcal{L}_e +\beta *\mathcal{L}_{f}+ \gamma *\mathcal{L}_{me}.
    \label{e18}
\end{equation}

Specifically, we follow the strategy of weighted sum in ADAGCN to obtain $\mathcal{L}_{semi}$. In addition, $\beta$ and $\gamma $ is utilized as a learnable parameter to coordinate the all objectives. The detailed learning procedure of the proposed GANN is shown in Algorithm \ref{ALGORITHM}.

\section{4\hspace{0.3cm}Experiments}
\subsection{Experimental Setup}
\textbf{Datasets.} To demonstrate the effectiveness of the proposed GANN, we conduct extensive experiments on eight datasets, including DBLP \footnote{{\footnotesize{}{}\texttt{\footnotesize{}https://dblp.uni-trier.de}{\footnotesize{}}}}, ACM \footnote{{\footnotesize{}{}\texttt{\footnotesize{}https://dl.acm.org/}{\footnotesize{}}}}, AMAP\cite{ama}, AMAC\cite{ama}, Cora-ML\cite{37}, CiteSeer\cite{38}, PubMed\cite{39}, and MS-Academic\cite{40}. 

DBLP is a network structure that connects several authors from diverse domains, including machine learning, database, etc. The edge connection indicates whether they are co-authors. ACM is a network of papers, and an edge connection is created when the same author is responsible for multiple publications. AMAP is a graph of Amazon product purchases, where nodes represent products and edges indicate if they are frequently purchased together. AMAC is comparable to AMAP and is included in Amazon's graph of often purchased items. The last four of them are text classification datasets. Cora-ML, CiteSeer and PubMed are citation graphs. In other words, each node inside the network represents an article, while the edge connecting nodes denotes their citation relationship. The edge link between network nodes in the MS-Academic dataset shows co-authorship. 
Besides, Table \ref{data} summarizes the data statistics.

\begin{table}[htbp]
    \centering{
    \resizebox{\linewidth}{!}{
    \begin{tabular}{cccccc}
     \hline
    \textbf{Dataset} & \textbf{Nodes} & \textbf{Edges} & \textbf{Features} & \textbf{Classes} & \textbf{Label Rates} \\
    \hline
    DBLP & 4507 & 7056 & 334 & 4 & 0.018 \\
    ACM & 3025 & 26 256 & 1870 & 3 & 0.019 \\
    AMAP & 7650 & 287 326 & 745 & 8 & 0.020 \\
    AMAC & 13 752 & 491 722 & 767 & 10 & 0.014 \\
    Cora-ML & 2810 & 7981 &2879 & 7  & 0.047 \\
    CiteSeer & 2110 & 3668 &3703 & 6  &0.036  \\
    PubMed & 19 717 & 44 324 &500 & 3  &0.003 \\
    MS-Academic & 18 333 & 81 894 &6805 & 15 &0.016 \\
    \hline 
    \end{tabular}}}
    \caption{Dateset Statistics}
    \label{data}
\end{table}

\textbf{Baselines.}
We choose thirteen representative models from the relevant directions, and they can be categorized into three types. Concretely, the first type works with graph neural networks: GCN (with early stopping), V.GCN\cite{1}, GCN-Cheby\cite{c-gcn} and GAT\cite{19}. Besides, the second type focuses on specific feature usage: FAGCN\cite{8}, SGC\cite{sgc} and MixupForGraph\cite{mixup}. Moreover, the third type focuses on the use of structure: JK-Net\cite{43}, PPNP, APPNP\cite{7}, GWNN\cite{gwnn}, GPRGNN\cite{gprgnn} and ADAGCN\cite{11}.

\begin{table*}[t]
\centering{
\resizebox{\textwidth}{!}{
\begin{tabular}{ccccccccc}
\hline
\textbf{Model} & \textbf{DBLP}       & \textbf{ACM}        & \textbf{AMAP}       & \textbf{AMAC}       & \textbf{Cora-ML}   & \textbf{CiteSeer}   & \textbf{PubMed}     & \textbf{MS-Academic} \\
\hline
GCN            & 78.79±0.54          & 91.85±0.40          & 81.15±1.05          & 58.96±0.96          & 82.13±0.72          & 74.42±0.73          & 76.89±0.54          & 92.01±0.08            \\
V.GCN          & 78.32±0.78          & 91.06±0.44          & 81.41±1.47          & 58.71±0.76          & 82.61±0.76          & 74.07±0.69          & 76.70±0.63          & 91.83±0.16            \\
GCN-Cheby      & 80.33±0.97          & 91.01±0.56          & 91.56±0.75          & 77.31±1.18          & 82.97±0.67          & 72.21±0.54          & 75.01±0.74          & OOM                   \\
GAT            & 79.97±0.58          & 90.16±0.52          & 92.38±0.14          & 79.04±0.58          & 83.53±0.16          & 72.17±0.73          & 77.85±0.26          & 89.47±0.20            \\
FAGCN          & 79.98±0.89          & 90.38±0.86          & 90.06±1.04          & 80.41±0.62          & 83.72±0.64          & 71.97±1.05          & 77.00±0.84          & 90.53±0.20            \\
SGC            & 79.27±1.07          & 91.75±0.48          & 90.17±0.75          & 77.48±0.79          & 83.65±1.16          & 73.16±1.07          & 79.30±0.60          & 89.79±0.83            \\
MixupForGraph  & 74.29±0.71          & 89.33±0.59          & 87.28±1.08          & 59.65±0.71          & 78.76±0.74          & 68.68±0.91          & 72.70±0.40          & 85.59±1.26            \\
JK-Net         & 79.62±0.58          & 90.56±0.56          & 91.68±0.82          & 75.95±1.07          & 82.93±0.38          & 72.78±0.44          & 76.54±0.57          & 89.08±0.05            \\
PPNP           & 80.40±0.42          & 91.64±0.44          & 86.27±0.74          & 66.15±0.74          & 85.29±0.30          & 75.42±0.27          & OOM                 & OOM                   \\
APPNP          & 79.62±0.74          & 91.54±0.34          & 86.02±0.76          & 64.99±1.25          & 85.09±0.25          & 75.51±0.42          & 78.63±0.65          & 92.12±0.09            \\
GWNN           & 79.27±0.54          & 90.91±0.49          & 90.75±1.26          & 75.84±0.87          & 83.84±0.55          & 74.06±0.80          & 78.69±0.94          & 90.27±0.53            \\
GPRGNN         & \underline {80.92±0.20}    & 91.35±0.35          & \underline {93.08±0.26}    & \underline {80.71±0.33}    & 85.07±0.24          & 74.09±0.41          & 77.38±0.48          & 90.47±0.18            \\
ADAGCN         & 80.45±0.83          & \underline {92.04±0.62}    & 85.13±0.62          & 67.58±0.99          & \underline {85.67±0.59}    & \underline {76.24±0.43}    & \underline {79.38±0.63}    & \underline {93.11±0.23}      \\
\hline
GANN(Ours)           & \textbf{82.04±0.22} & \textbf{93.23±0.34} & \textbf{94.05±0.37} & \textbf{81.83±0.24} & \textbf{87.12±0.29} & \textbf{77.94±0.32} & \textbf{81.09±0.23} & \textbf{94.16±0.18} \\
\hline
\end{tabular}}}
\caption{This is the result of using $20$ samples per cluster for all datasets. We evaluate the node classification tasks using the average accuracy($\%$) and standard deviation as a criterion. Experiments use $10$ random seeds. OOM denotes “out of memory”. Bolded results are optimal, and underlined results are suboptimal.}
\label{res1}
\end{table*}

\textbf{Training.} 
In this part we present the data and parameter details used for the experiments. 1) For data usage, we employ three scales for the splitting of the data training set. The first method utilizes $20$ training samples per cluster across all datasets; the results are displayed in Table \ref{res1}. The second strategy employs $10$ for AMAC, and $15$ for the remaining datasets. Table \ref{res2} shows the results. And the third method uses $7$ samples per cluster for all datasets. Table \ref{res3} details the outcomes. The validation set has 500 samples, whereas the test set contains the remaining samples. 
2) For parameters using, a combination of parameters with lr of $0.01$, $topk$ of $10$, hidden size of 5000 and $\lambda$ of $1$ is used on all eight datasets. We set the appropriate number of layers for Cora-ML, CiteSeer, PubMed to $12$, $9$, $11$, and other datasets to $5$. For PubMed and MS-Academic, we set the dropout value to $0.2$, and for the other datasets, we set it to $0$. For PubMed and MS-Academic, we use 0.7 and 0.1, 0.5 and 3 for $\gamma$ and $\beta$, and 0.5 and 0.5 for other datasets. For AMAC, CiteSeer and Ms-Academic, we set weight decay to $1e-6$, $1e-3$ and $1e-5$, and for the other datasets, we set it to $1e-4$. Throughout the experiment, we use $pytorch3$ for code development. We use an NVIDIA GeForce RTX 3080 card with 64GB of RAM and a 20-core CPU for all datasets. We compare model outcomes by tweaking baseline source code to best fit their original parameters. The validation set's best model predicts the test set. We select category accuracy as the statistic for evaluation, and 10 random seeds are used for all experimental findings.

\begin{table*}[h!]
\centering{
\resizebox{\textwidth}{!}{
\begin{tabular}{ccccccccc}
\hline
\textbf{Model} & \textbf{DBLP}       & \textbf{ACM}        & \textbf{AMAP}       & \textbf{AMAC}       & \textbf{Cora-ML}   & \textbf{CiteSeer}   & \textbf{PubMed}     & \textbf{MS-Academic} \\
\hline
GCN            & 77.86±0.78          & 90.81±0.99          & 79.79±0.71          & 56.80±1.57          & 81.93±1.39          & 72.09±0.80          & 75.65±1.22          & 90.80±0.78            \\
V.GCN          & 77.50±1.15          & 90.60±0.30          & 79.54 ± 1.01        & 56.49±0.74          & 81.91±1.16          & 72.54±1.07          & 75.76±1.15          & 90.77±0.86            \\
GCN-Cheby      & 77.96±0.82          & 90.90±1.01          & 90.22±1.16          & 75.43±0.61          & 80.80±0.47          & 71.39±1.17          & 75.55±0.22          & OOM                   \\
GAT            & 78.43±0.23          & 90.01±0.95          & 92.36±0.50          & 78.98±0.41          & 81.31±0.29          & 71.94±1.12          & \underline{78.05±0.46}          & 88.23±0.48            \\
FAGCN          & 79.74±0.83          & 90.26±0.68          & 89.74±1.26          & \textbf{80.65±0.95} & 83.28±0.90          & 69.84±1.43          & 76.31±1.51          & 90.22±0.83            \\
SGC            & 79.21±1.37          & 91.11±0.89          & 89.87±1.28          & 76.66±0.96          & 82.90±0.57          & 72.23±0.84          & 77.90±0.45          & 89.37±1.04            \\
MixupForGraph  & 73.22±0.66          & 87.85±1.09          & 86.51±1.13          & 58.15±1.27          & 77.50±1.10          & 68.69±1.09          & 72.60±0.85          & 84.03±1.52            \\
JK-Net         & 78.91±0.51          & 90.39±0.40          & 91.50±0.41          & 74.11±0.96          & 80.74±1.02          & 73.07±0.54          & 76.11±0.67          & 89.04±0.21            \\
PPNP           & 80.09±1.07          & 91.29±0.46          & 84.37±1.22          & 65.66±1.02          & 82.58±1.03          & 75.41±0.63          & OOM                 & OOM                   \\
APPNP          & 79.18±0.54          & 91.34±0.64          & 85.60±0.50          & 65.22±1.18          & 82.90±1.18          & 75.37±0.93          & 76.98±1.61          & 92.08±0.77            \\
GWNN           & 78.08±1.18          & 91.01±0.40          & 89.99±1.37          & 75.09±1.13          & 81.85±0.97          & 72.13±0.82          & 76.76±1.36          & 90.20±0.84            \\
GPRGNN    & \underline{80.88±0.58}  & \underline{92.19±0.22} & \underline{92.70±0.11} & 80.13±0.72   & 84.14±0.31            & 73.68±0.27          & 77.62±0.24          & 90.30±0.22            \\
ADAGCN         & 79.38±0.83          & 92.09±0.64           & 85.39±0.98          & 61.83±1.50        & \underline{84.69±0.41}          & \underline{76.17±0.73}         &78.04±0.47              & \underline{92.66±0.17}      \\
\hline
GANN(Ours)  & \textbf{81.85±0.34} & \textbf{92.88±0.42} & \textbf{93.35±0.45} &\underline{80.27±0.52} & \textbf{86.33±0.58} & \textbf{76.78±0.62} & \textbf{80.45±0.33} & \textbf{93.38±0.14} \\
\hline
\end{tabular}}}
\caption{This is the result of using $10$ samples per cluster for AMAP and $15$ for others. We evaluate the node classification tasks using the average accuracy($\%$) and standard deviation as a criterion. Experiments use $10$ random seeds. OOM denotes “out of memory”. Bolded results are optimal, and underlined results are suboptimal.}
\label{res2}
\end{table*}

\begin{table*}[h!]
\centering{
\resizebox{\textwidth}{!}{
\begin{tabular}{ccccccccc}
\hline
\textbf{Model} & \textbf{DBLP}       & \textbf{ACM}        & \textbf{AMAP}   & \textbf{AMAC}    & \textbf{Cora-ML}   & \textbf{CiteSeer}   & \textbf{PubMed} & \textbf{MS-Academic} \\
\hline
GCN            & 76.39±1.69          & 89.38±1.53          & 79.65±0.97   & 54.32±0.98       & 79.96±0.86          & 69.34±0.76          & 71.83±1.46          & 89.30±1.12            \\
V.GCN          & 75.95±1.08          & 90.12±1.24          & 78.72±1.24    & 54.38±0.76      & 79.75±0.62          & 69.36±0.65          & 71.57±1.46          & 90.04±0.83            \\
GCN-Cheby      & 76.04±1.10          & 89.62±0.63          & 89.88±0.79    & 74.16±0.82      & 75.99±0.65          & 68.31±1.14          & 68.10±0.30          & OOM                   \\
GAT            & 77.33±0.31          & 89.66±0.78          & 90.77±0.75     & 77.21±0.56     & 78.67±0.45          & 70.86±0.97          & 68.94±0.28          & 87.86±0.66            \\
FAGCN          & 77.59±1.25          & 89.75±1.45          & 87.56±0.63    & 79.41±1.13      & 81.37±1.09          & 67.46±1.51          & 74.30±1.56          & 89.11±1.18            \\
SGC            & 77.30±1.79          & 90.46±0.52          & 88.33±1.69    & 73.21±1.68      & 79.18±1.02          & 66.15±1.80          & 72.82±1.67          & 88.36±0.61            \\
MixupForGraph  & 69.54±0.96          & 85.24±0.69          & 85.20±1.19    & 57.56±1.43      & 75.30±1.18          & 65.96±1.66          & 67.47±1.62          & 84.33±1.26            \\
JK-Net         & 78.08±1.08          & 89.94±0.56          & 89.42±0.86    & 72.45±1.22      & 77.72±0.71          & 70.65±0.23          & 67.92±0.31          & 88.32±0.25            \\
PPNP           & 79.91±1.32          & 90.90±0.40          & 83.26±1.59    & 64.89±0.95      & 81.01±1.47          & 72.38±1.05          & OOM                 & OOM                   \\
APPNP          & 79.38±1.42          & 90.49±1.19          & 83.67±1.18     & 65.19±1.04     & 80.94±1.16          & 72.93±1.50          & \underline{76.30±1.63}          & \underline{91.97±0.48}            \\
GWNN           & 75.53±1.07          & 90.45±0.59          & 88.92±1.22     & 74.32±0.78     & 80.83±1.08          & 70.63±0.93          & 74.50±0.82          & 89.84±0.32            \\
GPRGNN         & \textbf{80.83±0.65} & 90.68±0.19  & \underline{91.32±0.37}  & \underline{79.49±0.77}        & \underline{83.18±0.63}          & \underline{73.60±0.30}          & 69.58±0.19          & 88.90±0.22            \\
ADAGCN         & 77.61±1.78    & \underline{92.44±1.58}    & 84.73±0.50     & 59.48±1.32     & 82.08±0.99          & 72.95±1.38          & 74.17±1.11          & 91.81±0.49            \\
\hline
GANN(Ours)  & \underline{80.72±0.64} & \textbf{92.83±0.63} & \textbf{91.71±0.57} &\textbf{80.02±0.68} & \textbf{85.12±0.81} & \textbf{74.33±0.75} & \textbf{78.14±0.26} & \textbf{92.79±0.37} \\
\hline
\end{tabular}}}
\caption{This is the result of using $7$ samples per cluster for all datasets. We evaluate the node classification tasks using the average accuracy($\%$) and standard deviation as a criterion. Experiments use $10$ random seeds. OOM denotes “out of memory”. Bolded results are optimal, and underlined results are suboptimal.}
\label{res3}
\end{table*}

\begin{figure}[htbp]
% \small
    \centering
  \includegraphics[width=0.8\linewidth]{ 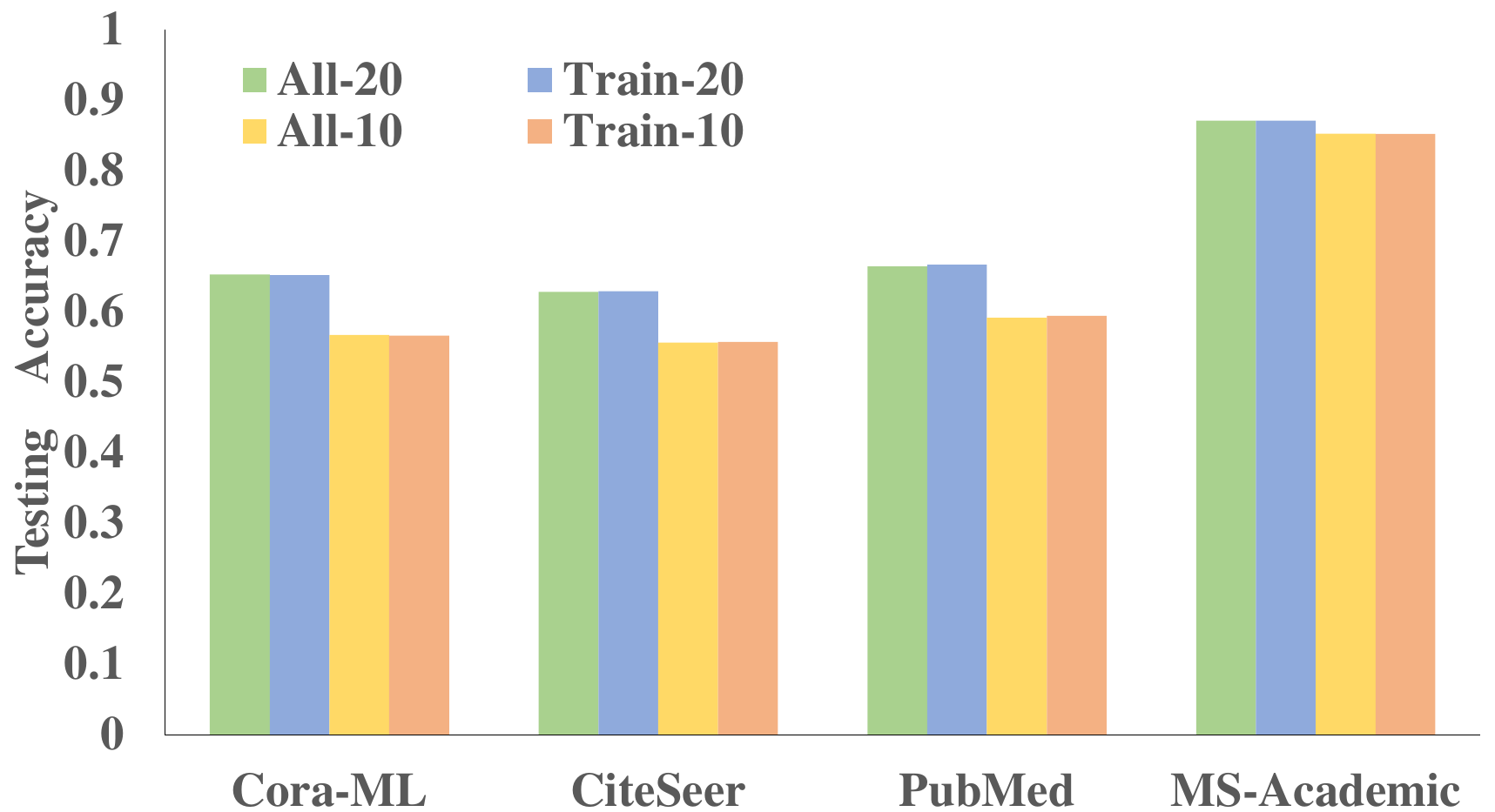}
  \caption{For experiments, only the attribute matrix is utilized. All-20 and All-10 indicate all input samples when the train size is 20 and 10; Train-20 and Train-10 indicate only 20 and 10 input samples. We give the average accuracy for ADAGCN run with 10 random seeds.}
  \label{figure2}
\end{figure}

\subsection{Performance Comparison}
As shown in Table \ref{res1}, Table \ref{res2} and Table \ref{res3}, we compare GANN with thirteen baselines at different labeling rates. From these results, we can infer the following conclusions. 1) First, in terms of the overall effect, our model GANN is able to maintain a stable accuracy and does not produce a large decrease in accuracy as the number of labels decreases. However, the accuracy of other models with superior performance, such as ADAGCN and GPRGNN, has decreased by $5\%$ to $13\%$. 2) GANN outperforms the majority of models in eight datasets, which indicates it can mine graph essential information and utilize graph attributes and structure information effectively. 

To further compare the model effects, we have selected a subset of models and datasets for visualization, as shown in Figure \ref{tsne} and Figure \ref{sim}. The illustration demonstrates that our proposed GANN model can generate high-quality node representations.

\begin{figure}[h!]
\begin{minipage}{0.48\linewidth}
\centerline{\includegraphics[width=1\textwidth]{ 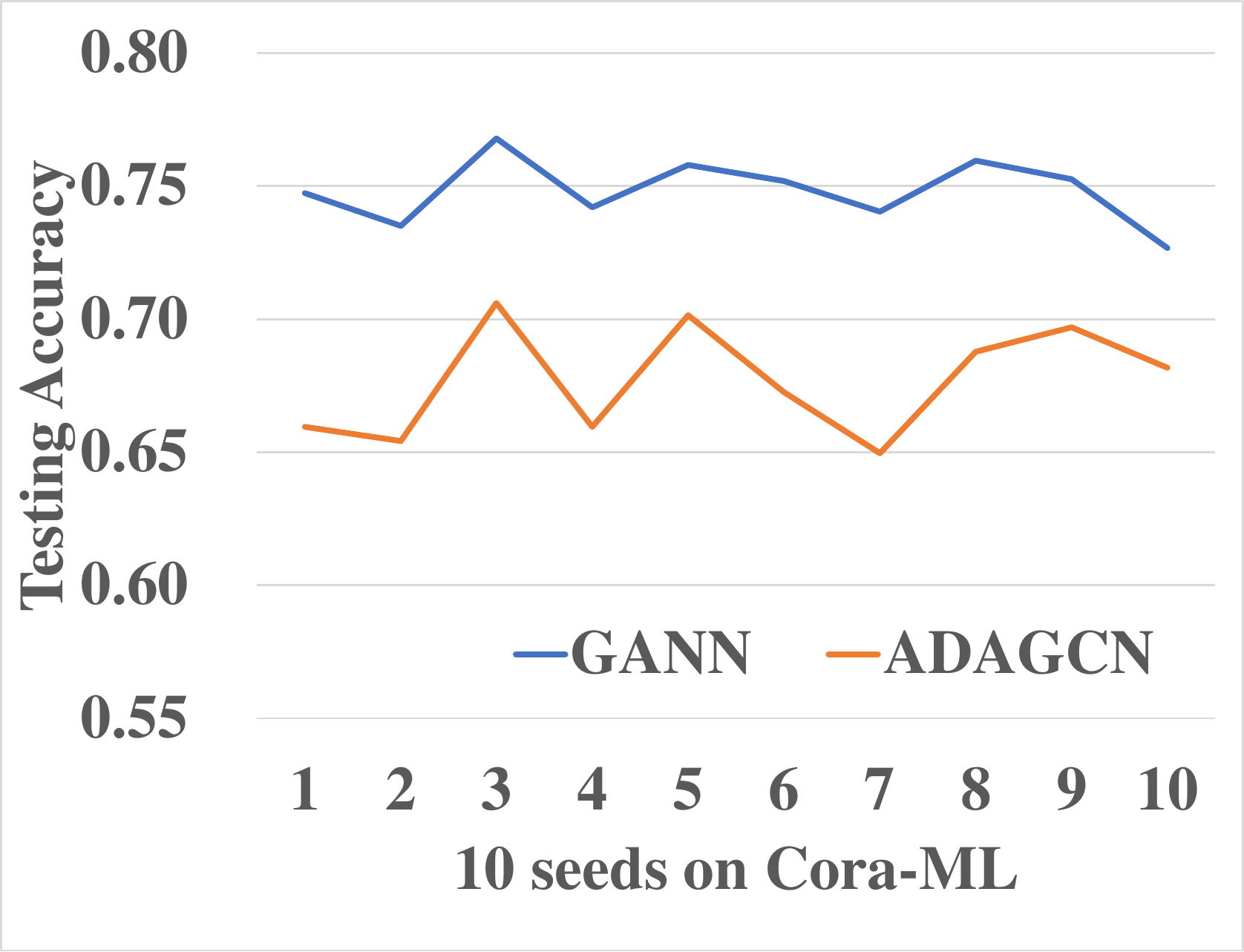}}
\centerline{(a) Cora-ML}
\vspace{3pt}
\end{minipage}
\begin{minipage}{0.48\linewidth}
\centerline{\includegraphics[width=1\textwidth]{ 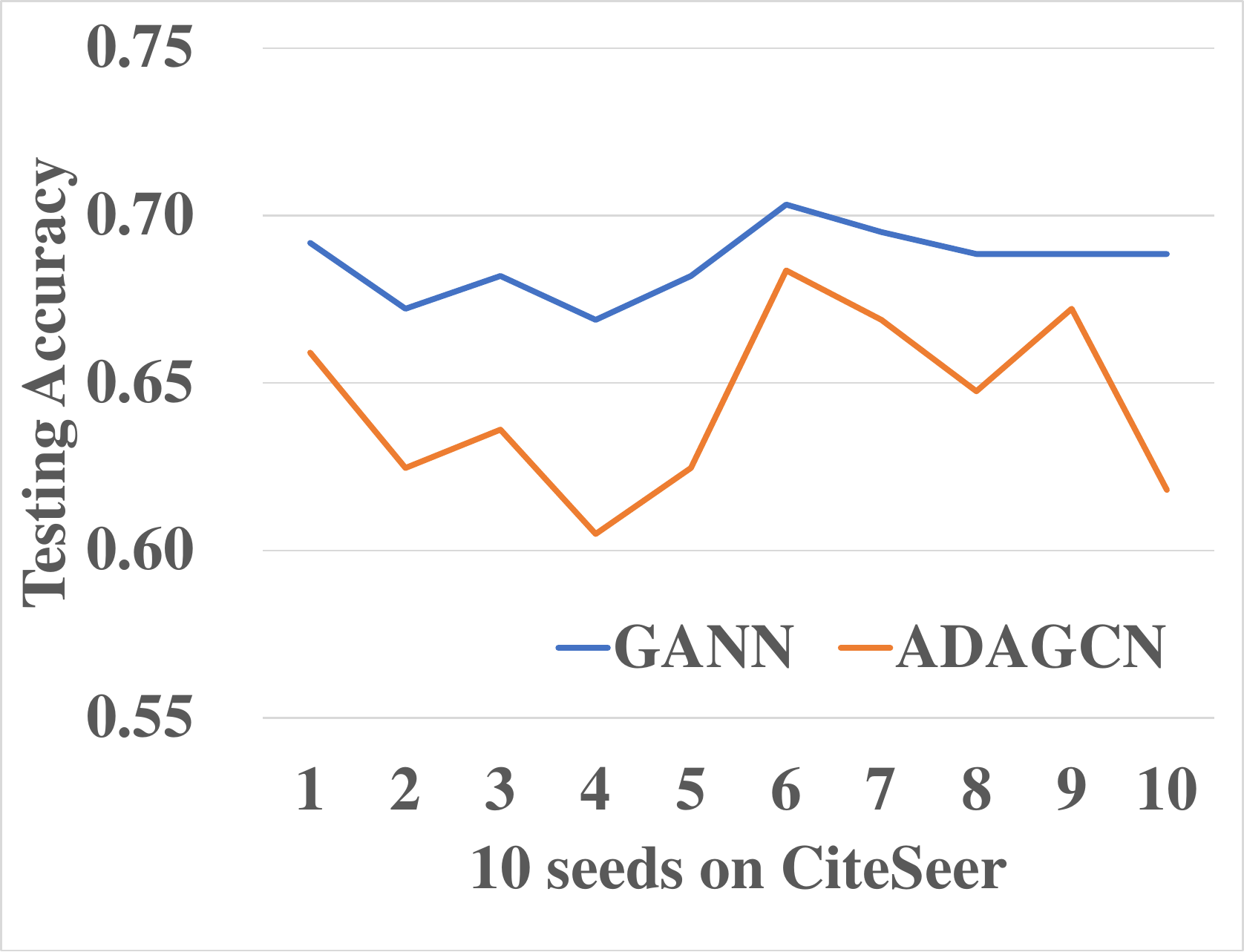}}
\centerline{(b) CiteSeer}
\vspace{3pt}
\end{minipage}
\caption{The accuracy of the test set with $10$ random seeds is reported.}
\label{figure3}
\end{figure}

\begin{figure}[h!]
\centering
\includegraphics[width=1\linewidth]{ 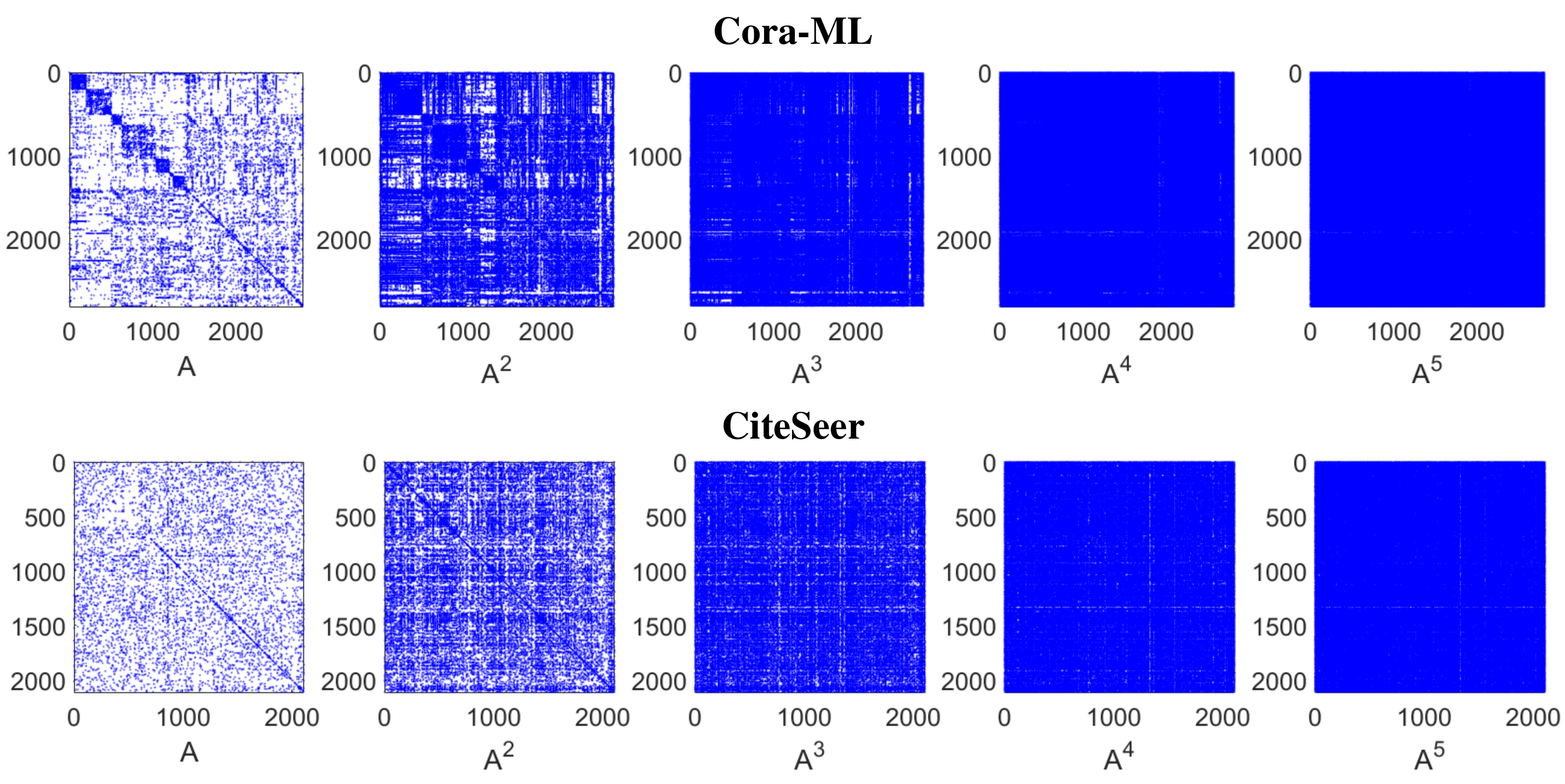}
\caption{From left to right, the one-hop to five-hop densities of the adjacency matrix are shown in order.}
\label{figure5}  
\end{figure}

\begin{figure}[htbp]
\begin{minipage}{0.31\linewidth}
\centering
\centerline{\includegraphics[width=1\textwidth]{ 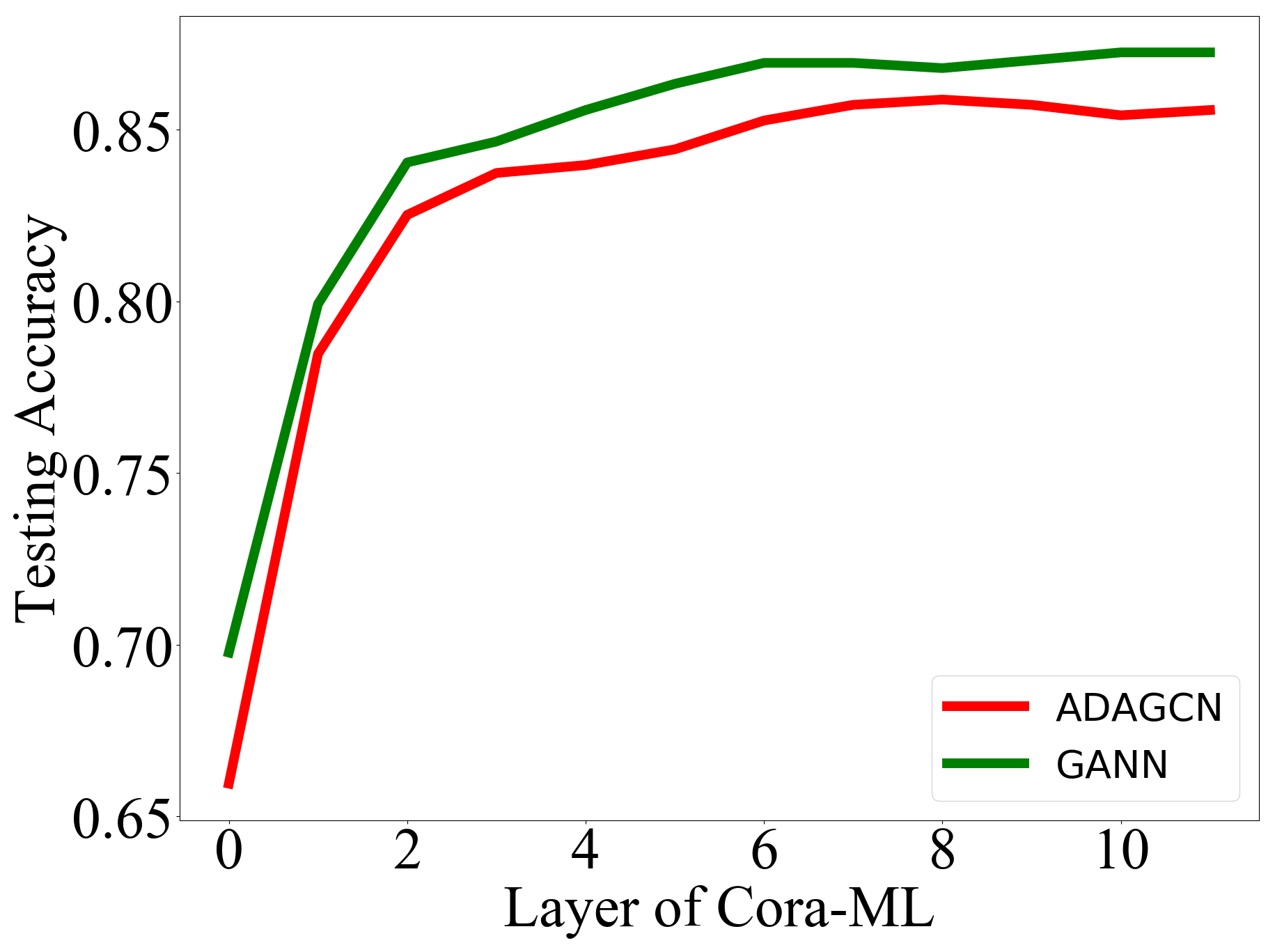}}
% \centerline{(a) test}
\vspace{1pt}
\end{minipage}
\begin{minipage}{0.31\linewidth}
\centering
\centerline{\includegraphics[width=1\textwidth]{ 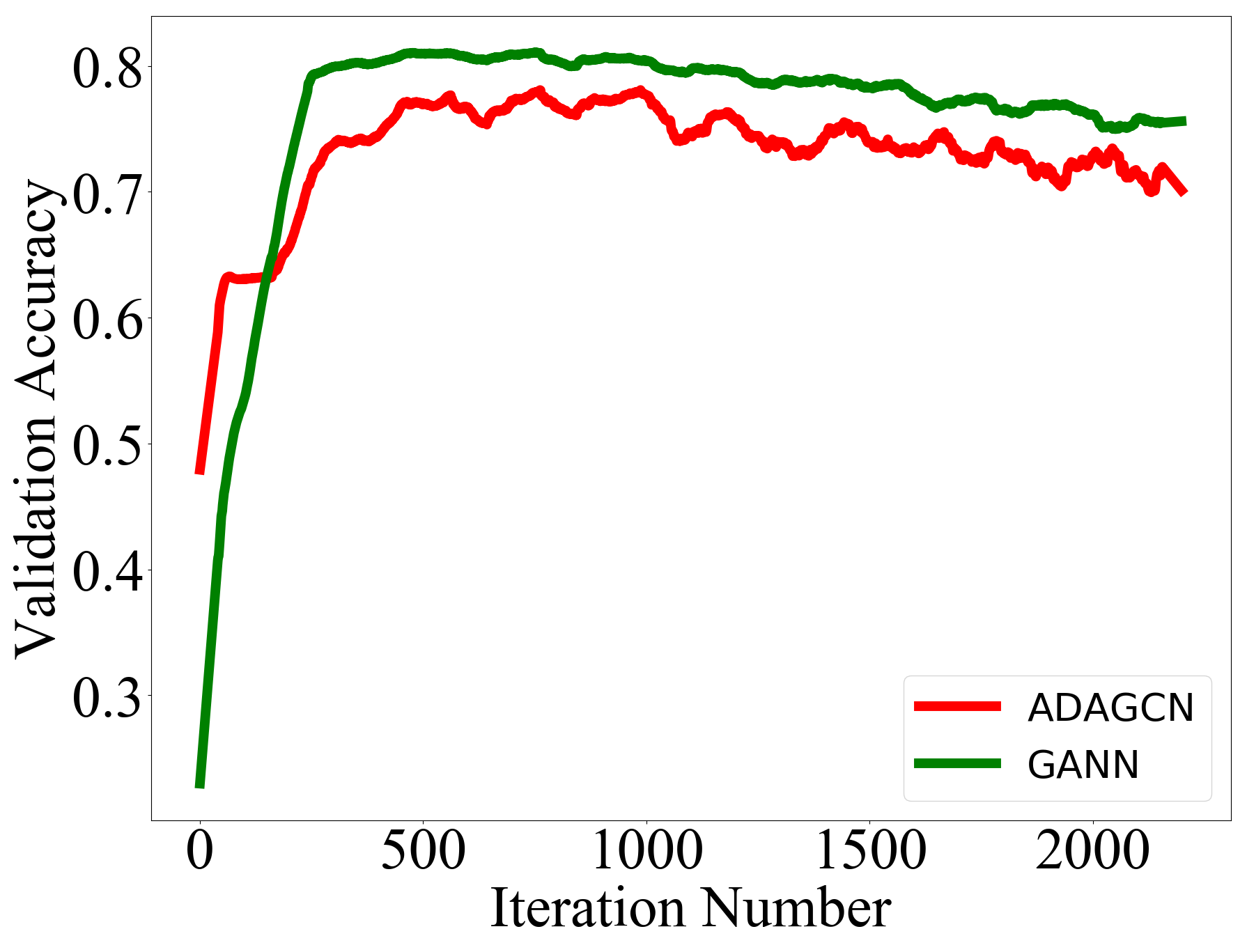}}
% \centerline{(a) Cora-ML}
\vspace{1pt}
\end{minipage}
\begin{minipage}{0.31\linewidth}
\centering
\centerline{\includegraphics[width=1\textwidth]{ 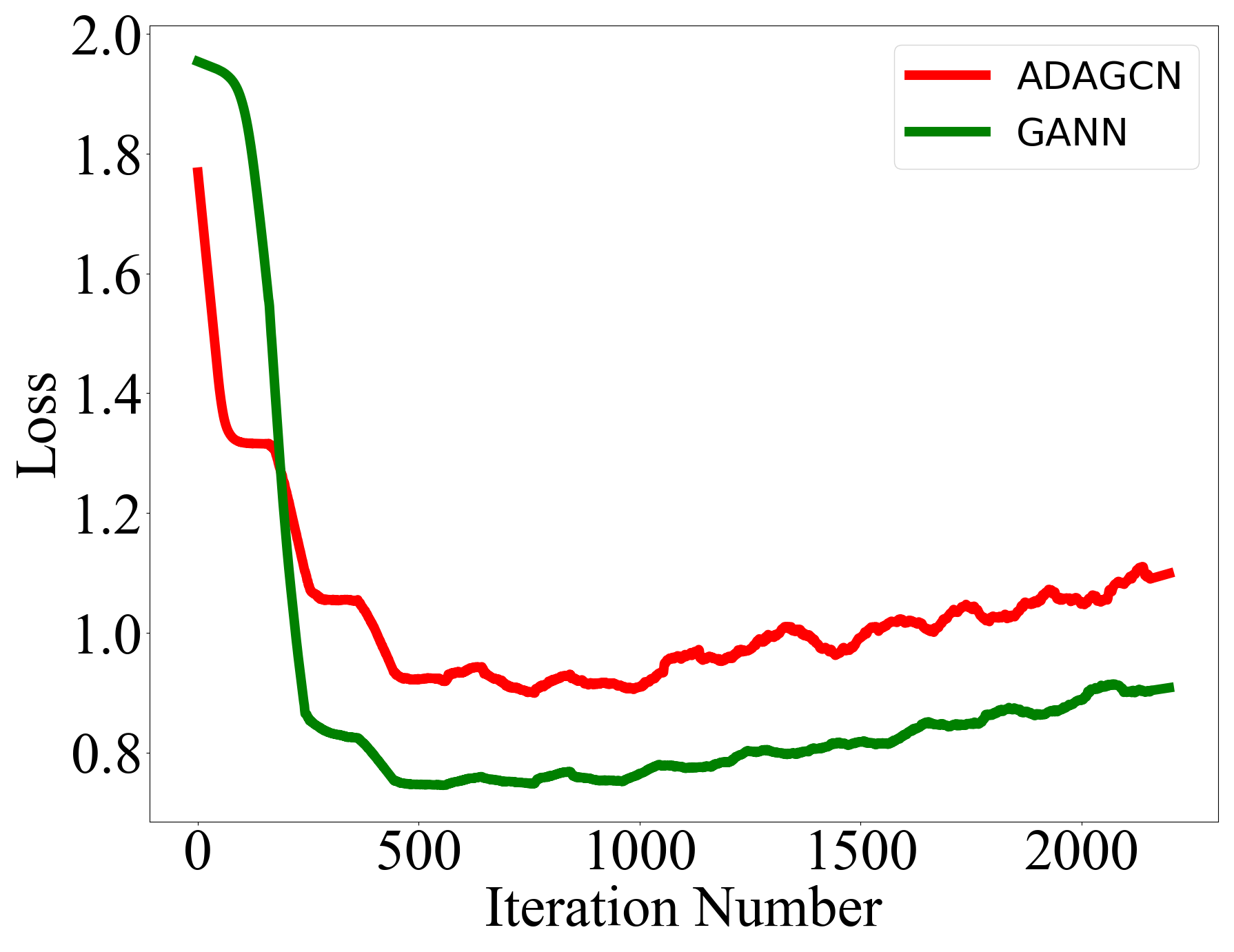}}
% \centerline{(b) CiteSeer}
\vspace{1pt}
\end{minipage}
\caption{Experiments are conducted on Cora-ML, and from left to right, the testing accuracy for various layers, validation set accuracy, and loss curve comparison plots are displayed.}
\label{figure6}  
\end{figure}

\begin{figure}[t]
\centering
\small
\begin{minipage}{0.49\linewidth}
\centerline{\includegraphics[width=1\textwidth]{ 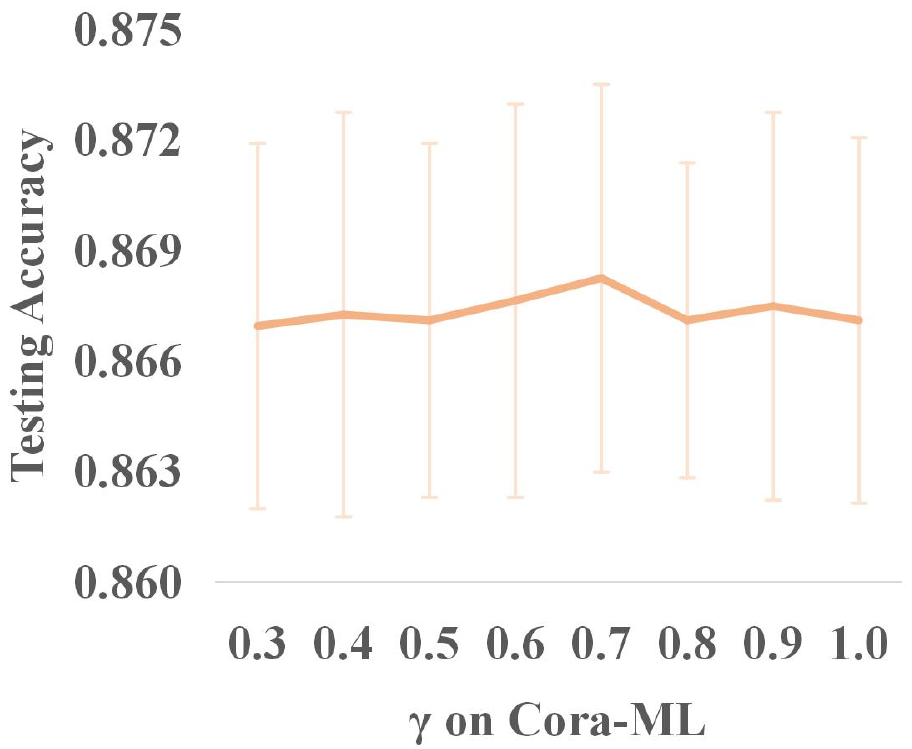}}
% \centerline{(a) $\beta$ on Cora-ML}
\vspace{1pt}
\centerline{\includegraphics[width=1\textwidth]{ 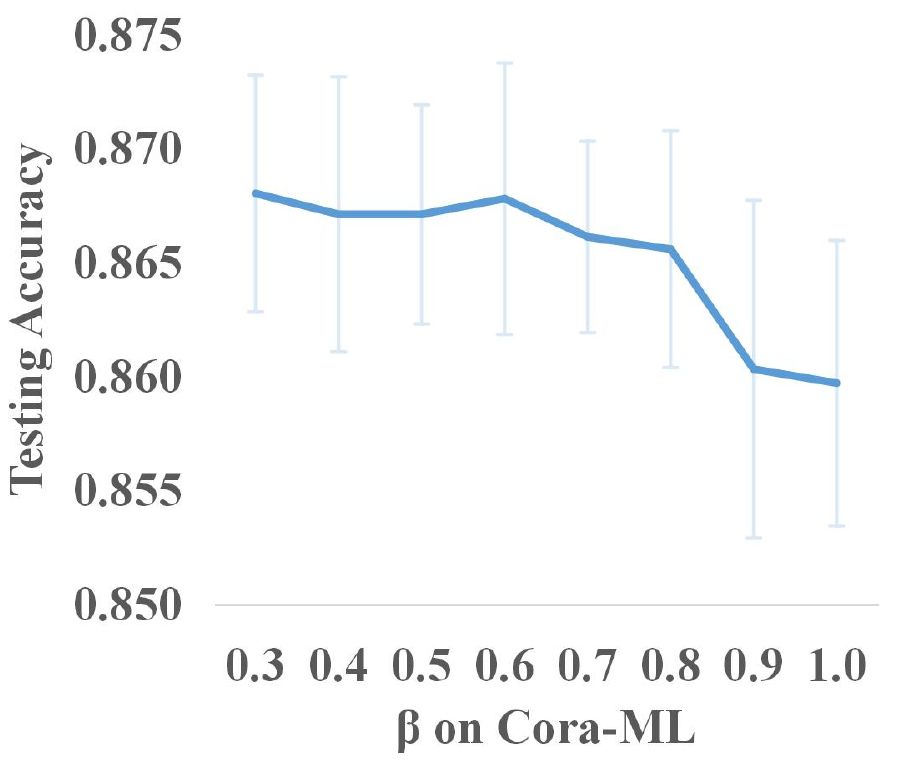}}
% \centerline{(c) $\gamma$ on CiteSeer}
\end{minipage}
\begin{minipage}{0.49\linewidth}
\centerline{\includegraphics[width=1\textwidth]{ 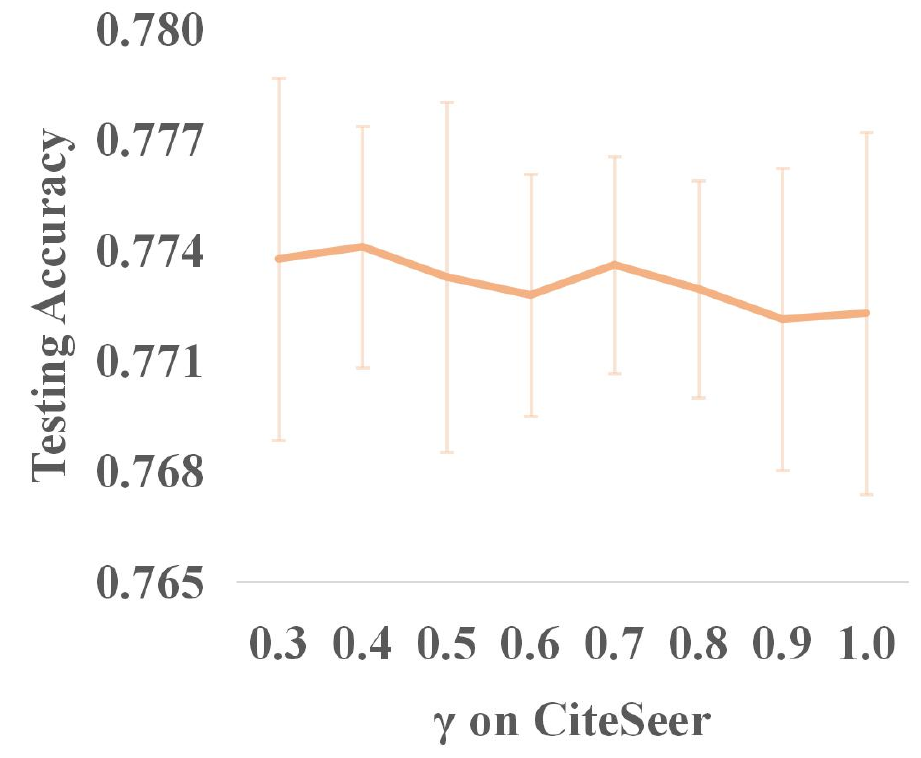}}
% \centerline{(b) $\beta$ on CiteSeer}
\vspace{1pt}
\centerline{\includegraphics[width=1\textwidth]{ 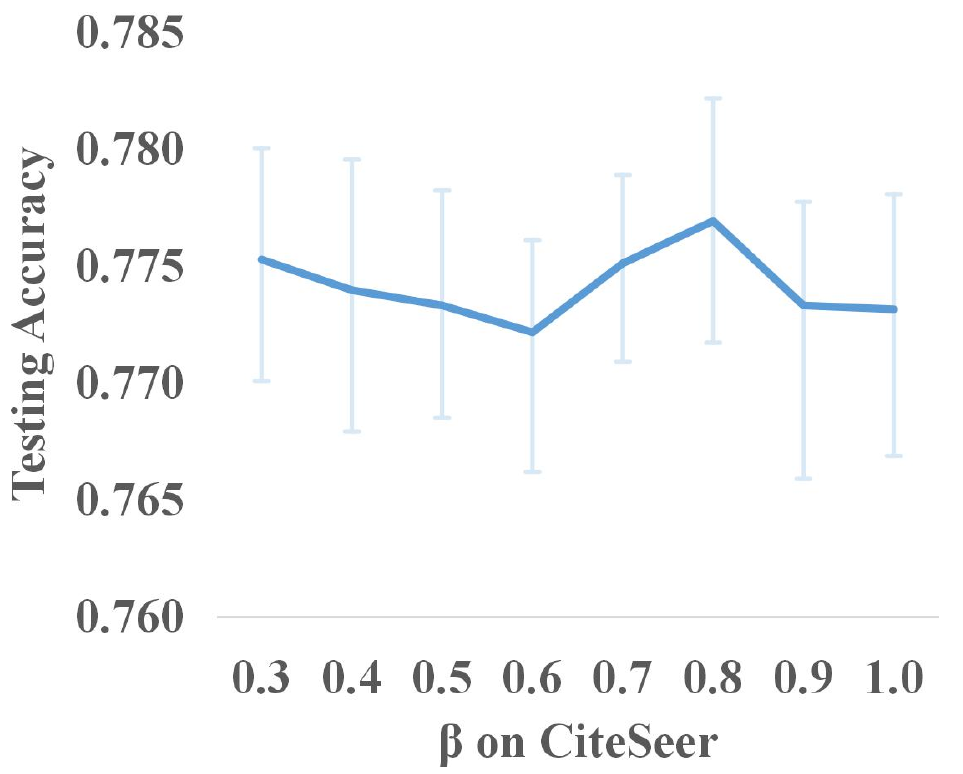}}
% \centerline{(d) $\gamma$ on CiteSeer}
\end{minipage}
\caption{For the sensitivity analysis of $\beta$ and $\gamma$, we give the average testing accuracy with $10$ random seeds and display the error lines using standard deviations.}
\label{figure7}
\end{figure}

\subsection{Ablation Study}
In this section, we do verification experiments on the previously introduced feature alignment rule, cluster center alignment rule, and the objective function.

\subsubsection{Analysis of the Feature Alignment Rule} ADAGCN's inefficient graph data exploitation is mentioned earlier. We execute a two-part experiment to illustrate that our feature alignment rule can utilize graph attributes. First, we show that only training set attributes can be utilized in the initial layer, regardless of whether the input is the whole attribute matrix. As demonstrated in Figure \ref{figure2}, the accuracy of using all samples is virtually the same as using only the training set in ADAGCN. Also, considering the number of the training set, it's plausible to believe that adding more labeled data enhances model outcomes. Then, we compare ADAGCN and GANN's first-layer testing accuracy to prove GANN's efficiency on Cora-ML and CiteSeer. Figure \ref{figure3} shows the results. The attribute information can be completely exploited during the training of GANN's first layer.

\subsubsection{Analysis of the Cluster Center Alignment Rule}
As indicated in ADAGCN limitations, the node representation created at deeper training layers tends to be oversmoothed. The explanation for this is that the outcomes of the adjacency matrix's higher order are no longer sparse. Figure \ref{figure5} shows that after several adjacency matrix multiplications, it is fully connected. Here in Figure \ref{figure6}, We display testing accuracy for varying numbers of layers, validation set accuracy, and loss curves for varying numbers of iterations on Cora-ML. Our results show that GANN improves deep layer training precision.

\begin{figure}[h!]
    \centering
    \includegraphics[width=0.6\linewidth]{ 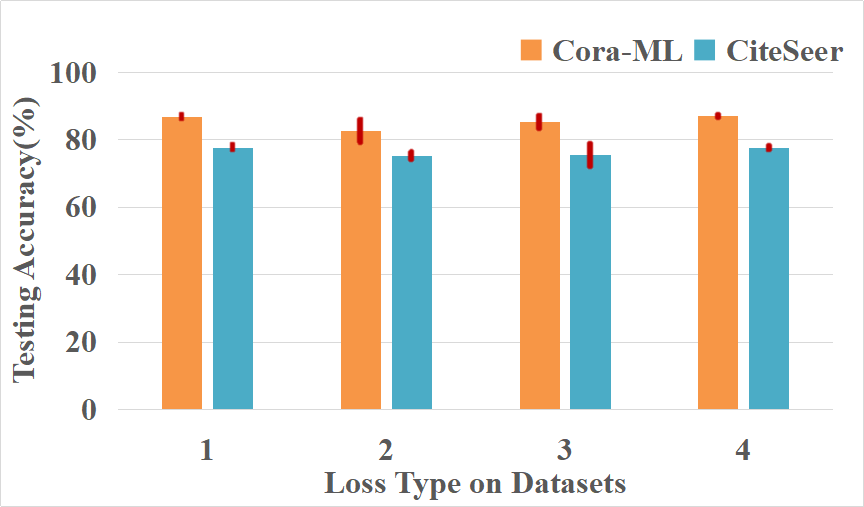}
    \caption{We report the average testing accuracy with 5 random seeds. Types 1, 2, 3, and 4 respectively indicate the implementation of the feature alignment rule, cluster center alignment rule, minimum entropy alignment rule, and GANN model.}
    \label{f1}
\end{figure}

\subsection{Analysis of the Objective Function}
In this subsection, we validate the loss in Eq.\eqref{e18} by evaluating the effects of three rules' objective functions on the model's output. We perform experiments on the Cora-ML and CiteSeer datasets and report the model's test accuracy and the standard deviation error lines. From Figure \ref{f1}, it is clear that the effect of the model using the feature alignment rule alone is comparable to the final performance of GANN. However, when the other two rules are used alone, the inaccuracy of the results is substantially bigger. But the best results achieved by the models corresponding to the three rules do not differ much. This highlights the necessity of considering the addition of regularization terms while individually applying rules.

\begin{figure}[h]
\centering
\begin{minipage}{0.48\linewidth}
\centerline{\includegraphics[width=1\textwidth]{ 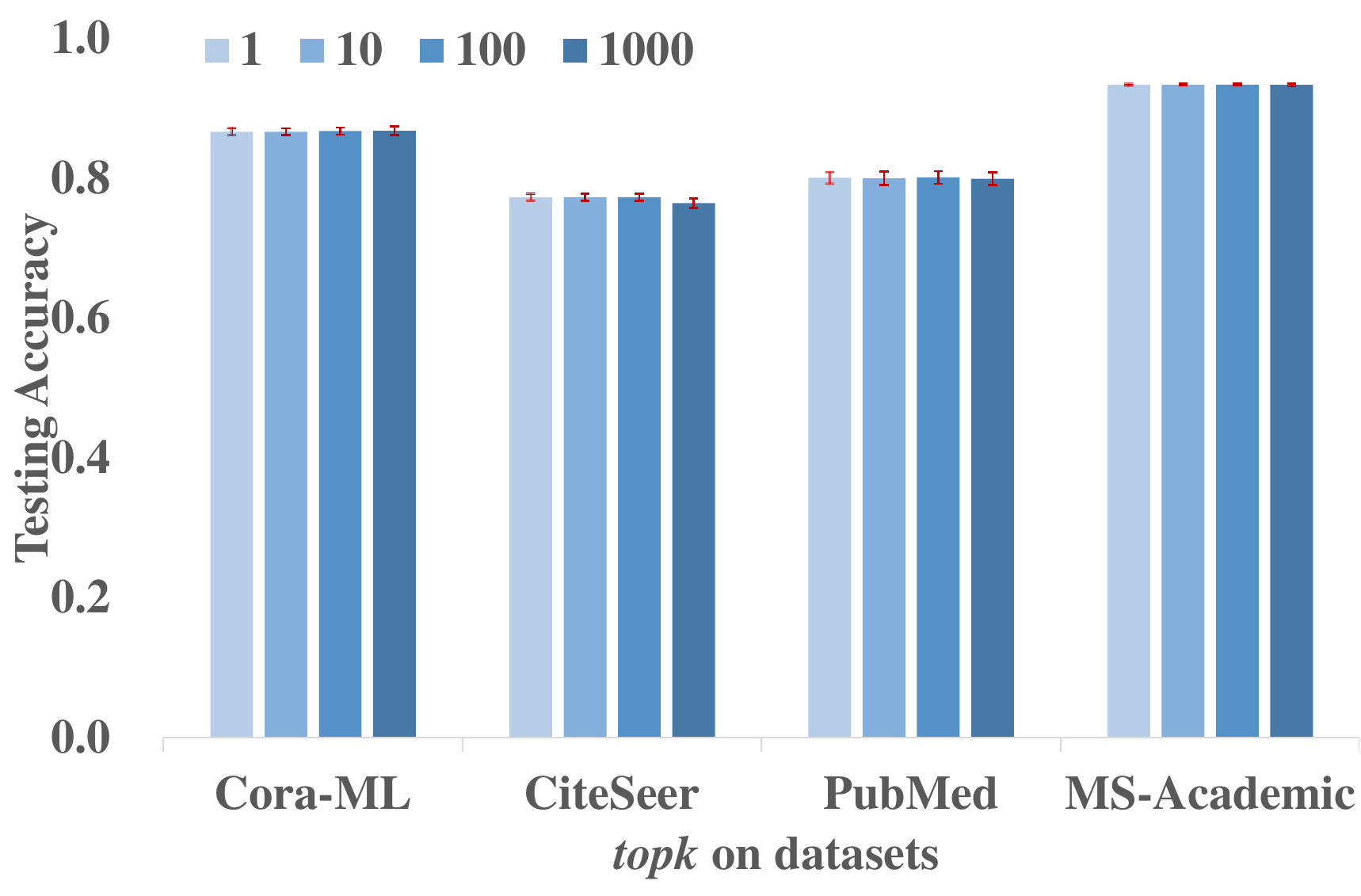}}
% \centerline{(a) $topk$}
\vspace{1pt}
\end{minipage}
\begin{minipage}{0.48\linewidth}
\centerline{\includegraphics[width=1\textwidth]{ 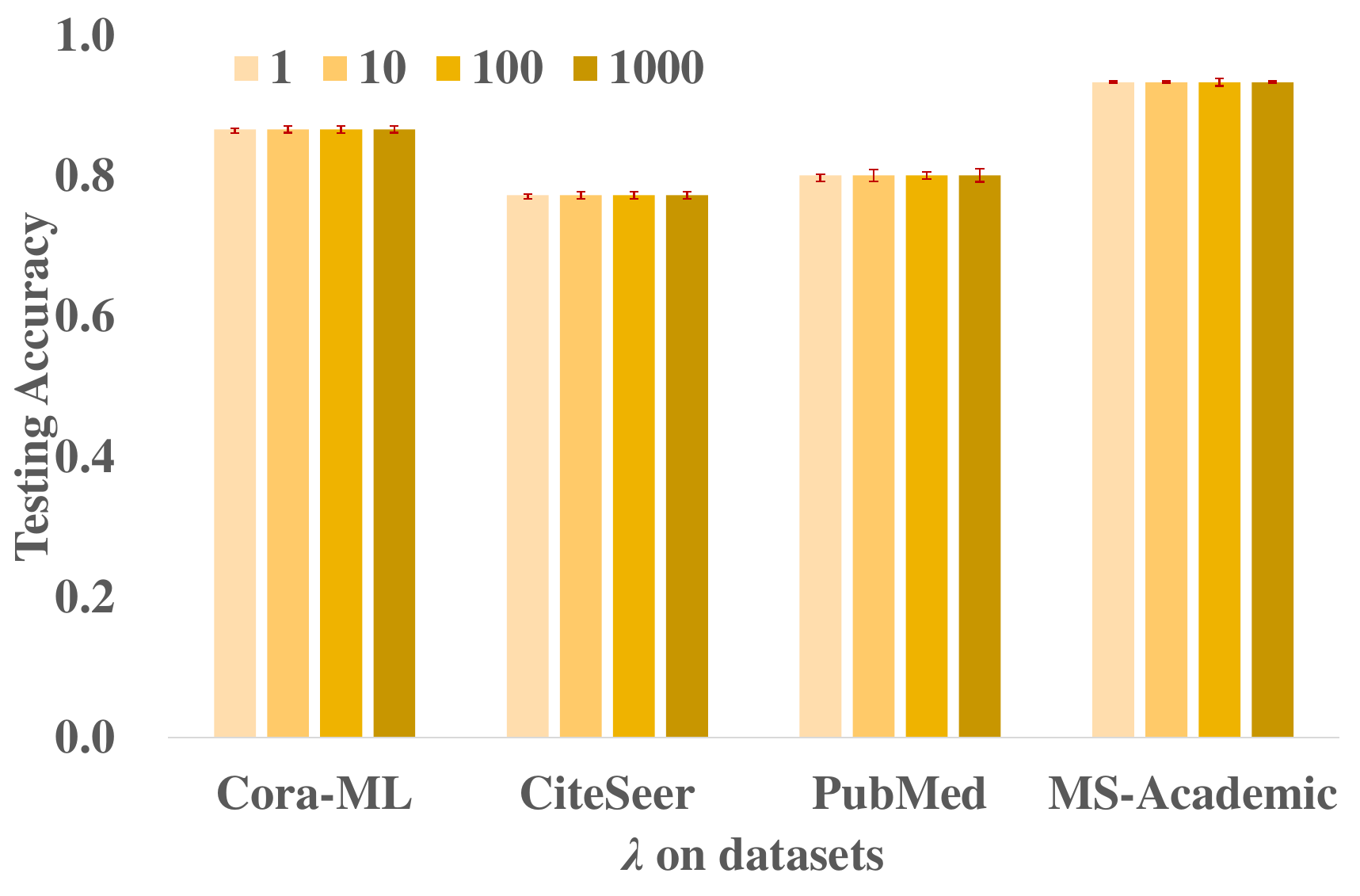}}
% \centerline{(b) }
\vspace{1pt}
\end{minipage}
\caption{Parameter sensitivity analysis of $topk$ and $\lambda$ in different datasets.}
\label{figure8}
\end{figure}

\begin{figure}[h!]
\centering
\small
\begin{minipage}{0.49\linewidth}
\centerline{\includegraphics[width=1\textwidth]{ 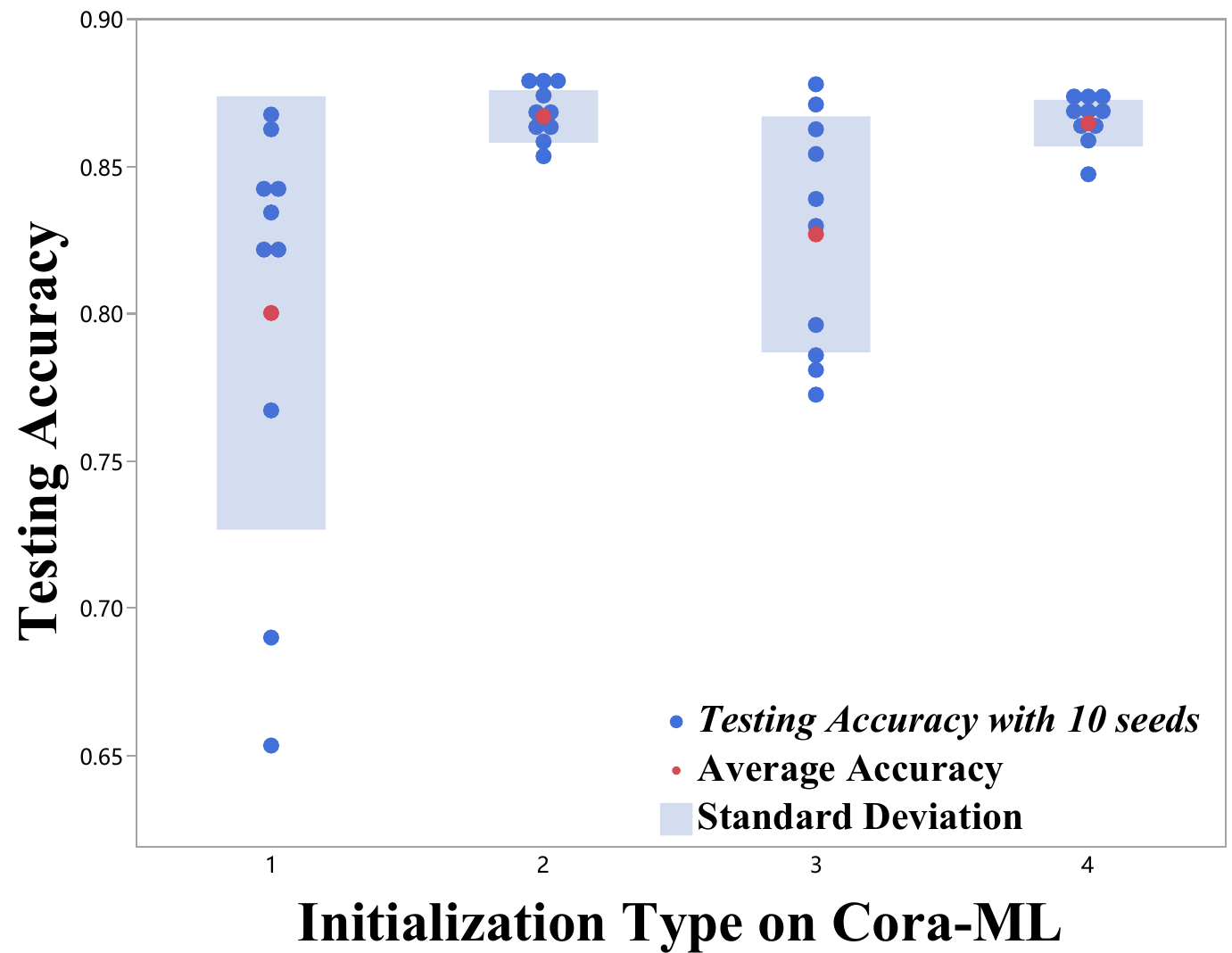}}
% \centerline{(a) Cora-ML}
\vspace{3pt}
\centerline{\includegraphics[width=1\textwidth]{ 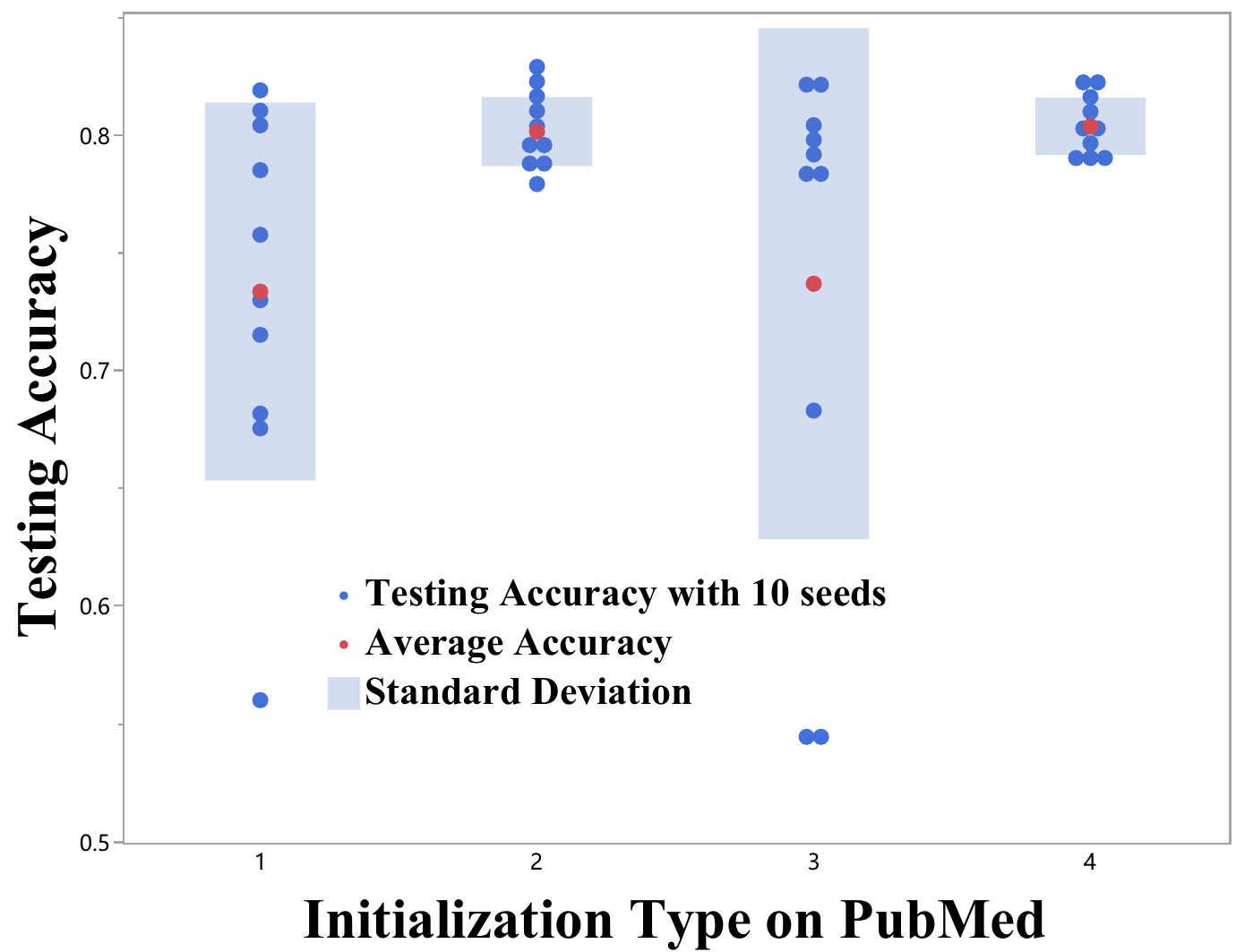}}
% \centerline{(c) PubMed}
\end{minipage}
\begin{minipage}{0.49\linewidth}
\centerline{\includegraphics[width=1\textwidth]{ 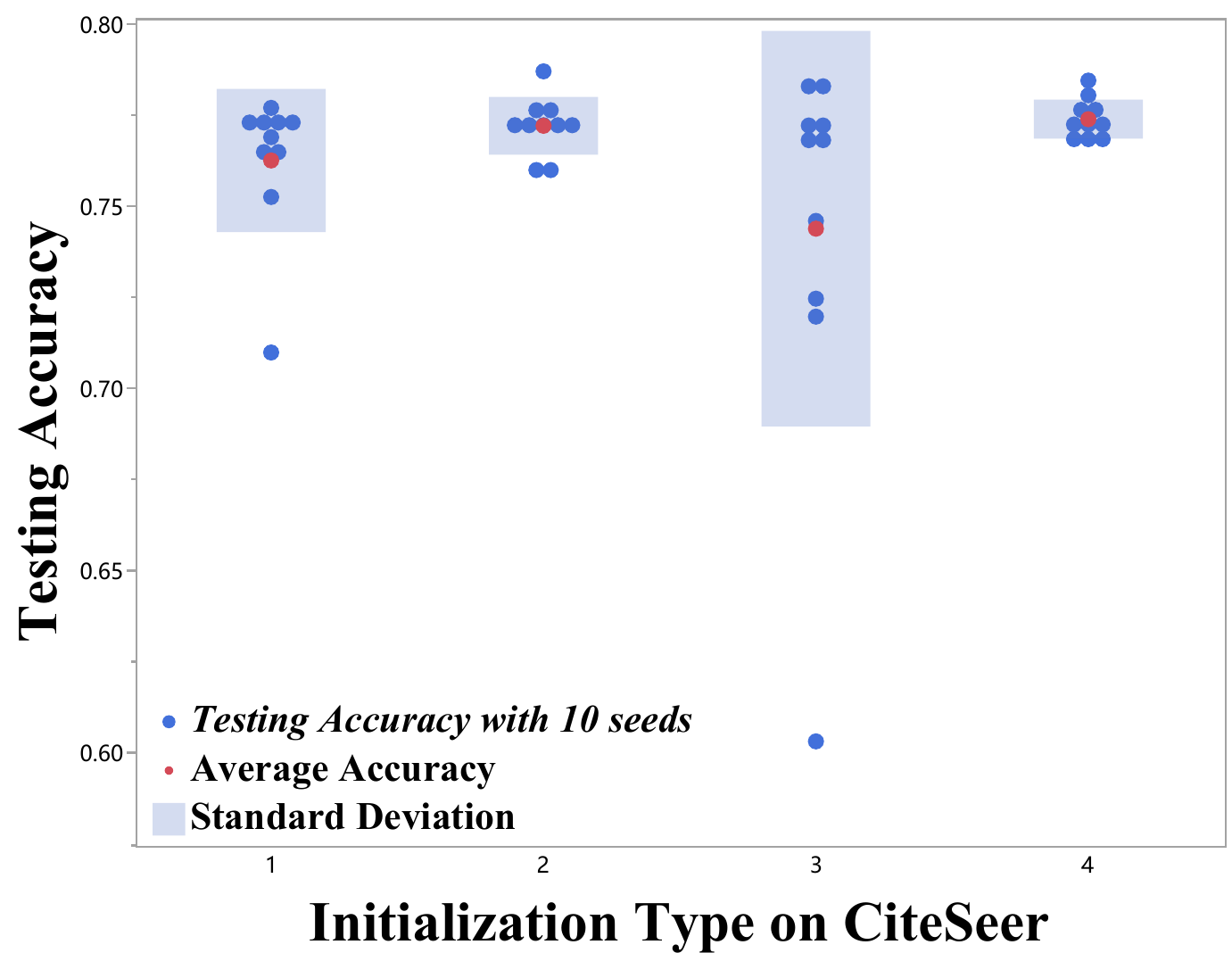}}
% \centerline{(b) CiteSeer}
\vspace{3pt}
\centerline{\includegraphics[width=1\textwidth]{ 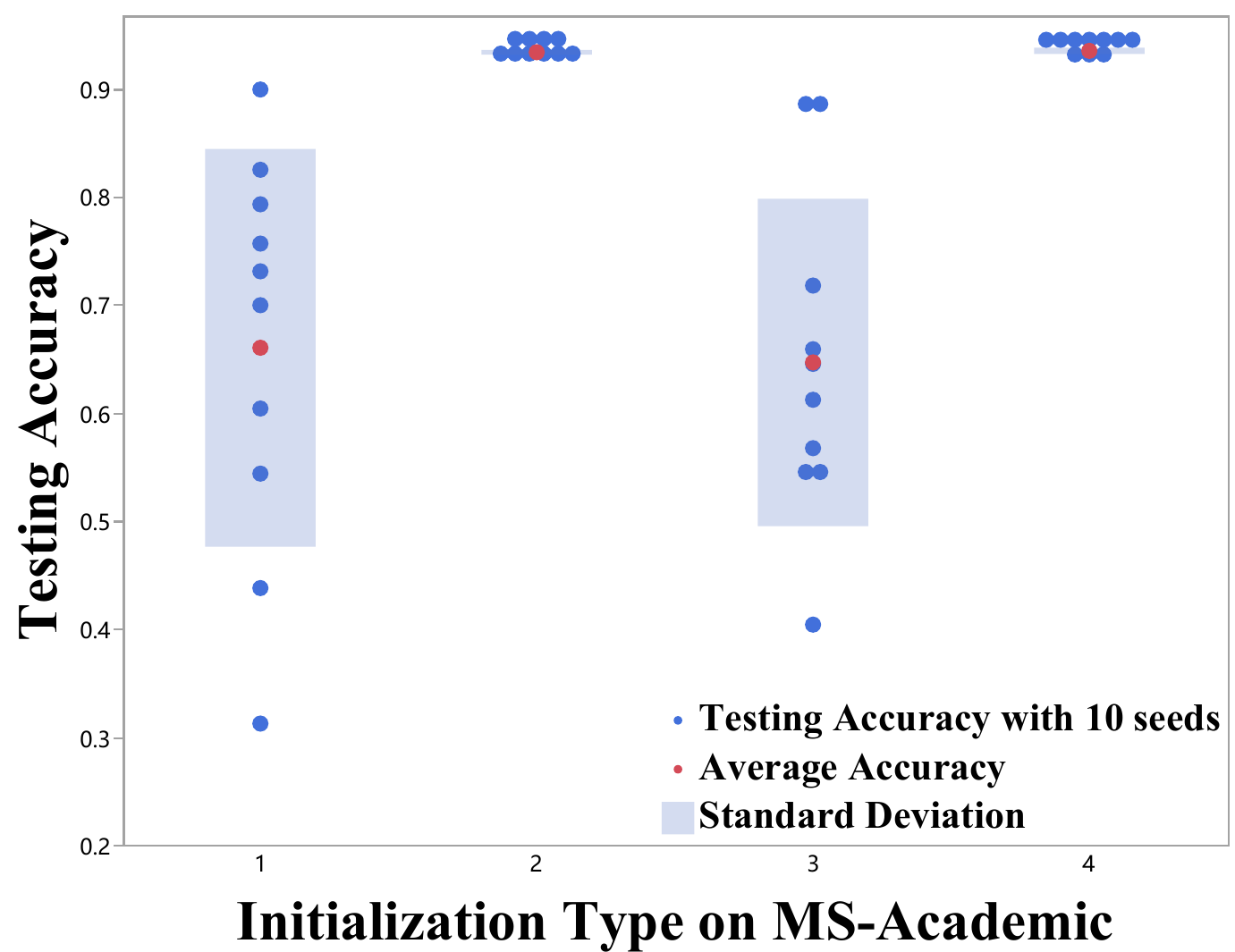}}
% \centerline{(d) MS-Academic}
\end{minipage}
\caption{We show the testing accuracy scatter plots for the bias modification of the MLP layer.}

\label{figure9}
\end{figure}

\subsection{Sensitivity Analysis}
In order to conduct an in-depth investigation of the model's parameters and structure, we conducted an experimental analysis of the model's key parameters and MLP structure in this subsection.

\subsubsection{Parameters Analysis} We investigate four critical model training parameters: the minimum entropy alignment rule loss factor $\gamma$, the feature alignment rule loss factor $\beta$, the cluster center alignment noise factor $\lambda$, and $topk$. We analyze $\gamma$ and $\beta$ on Cora-ML and CiteSeer with $0.3$ to $1.0$ parameter values. 
As shown in Figure \ref{figure7}, the optimal parameter values can be determined.

Also, we compare $topk$ and $\lambda $ parameter experiments in the ranges $[1, 10, 100, 1000]$. As shown in Figure \ref{figure8}, both parameters do not significantly affect the model's outcomes. We think the cause may be data standardization. 

Due to the fact that these two parameters are unconnected to the MLP layer that generates the embedding, parameter alterations have little effect on the final output. We also evaluate the MLP layer. Different topologies may result in substantial experimental variation.

\subsubsection{Model Structural Analysis} 
As indicated, biasing the MLP layer affects the model output. 
To investigate the influence of bias on model generation, we introduce bias into the two MLP layers of GANN. We used $10$ random seeds to conduct experiments on four datasets and we set types $1$, $2$, $3$, and $4$ to signify applying bias in the first and second layers, the first layer alone, the second layer only, and no bias.

Figure \ref{figure9} shows that adding bias to the first MLP layer has minimal influence on the model but helps the Cora-ML dataset. However, introducing bias to the second MLP layer causes a major deviation in the model result, even though the bias is initialized to zero. This is because the second MLP layer classifies the node embedding. Adding bias here will affect the final classification probability and model training negatively.

\begin{figure}[h!]
\centering
\includegraphics[width=1\linewidth]{ 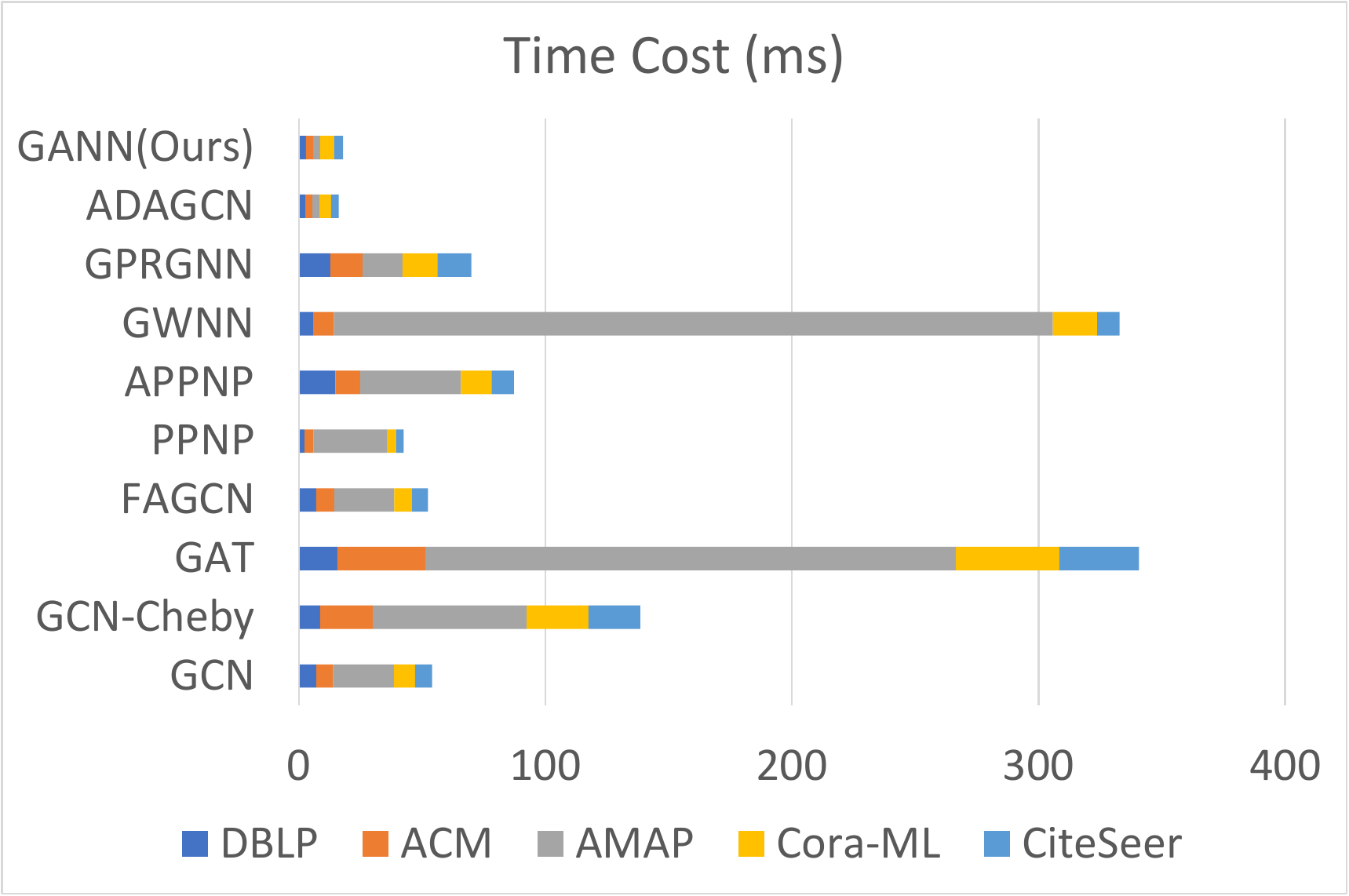}
\caption{The amount of time required by various models to complete one epoch of training under identical conditions. All model layers have two layers, and hidden layers have $128$ nodes.}
\label{time}  
\end{figure}

\subsection{Complexity Analysis}
In this section, we conduct analytical experiments on the model's time and space needs.

The results for time consumption are shown in Figure \ref{time}.  We run experiments utilizing 10 baselines on 5 datasets, each with 1000 epochs under the same conditions. We find that the time of GANN dominates the existing state-of-the-art methods. We analyze the root causes as follows: 1) Instead of using the inefficient and time-consuming GCN structure for data feature extraction, we directly multiply the adjacency matrix with the feature matrix as input, compute it just once, and store it for reuse. This significantly decreases the time required. 2) Two layers of simple MLP structure make the model structure simple and efficient to train. 

\begin{figure}[h!]
\centering
\includegraphics[width=1\linewidth]{ 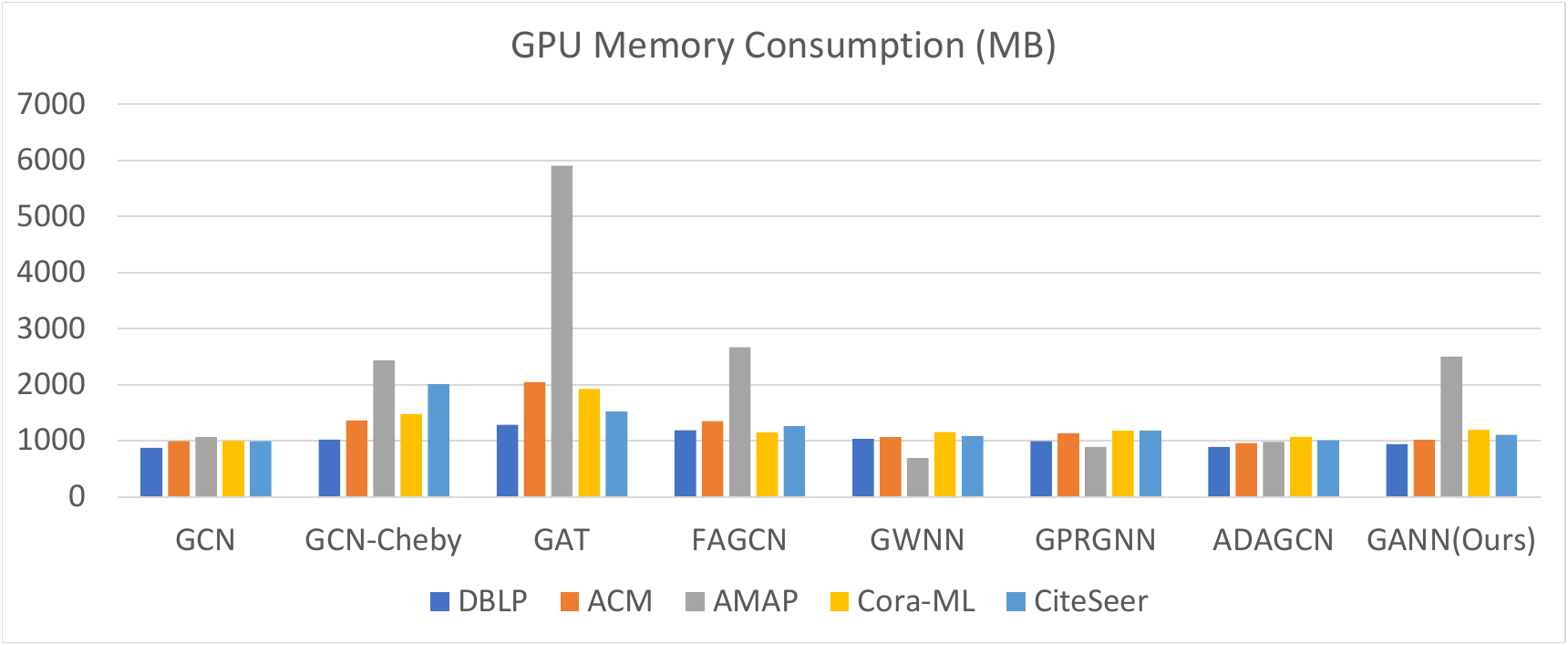}
\caption{GPU memory consumption of the models during training. All models use two layers and use $128$ hidden layer nodes.}
\label{gpu}  
\end{figure}

The GPU memory consumption results are shown in Figure \ref{gpu}. We employ eight baselines for trials on five datasets and record the maximum memory used for training models under the same conditions. According to the results, Our model GANN consumes GPU similarly to others, especially ADAGCN. Analysis: 1) GANN uses an incredibly simplistic model structure. 2) GANN iteratively trains between layers, removing the memory cost of layer extension.

\begin{figure}[h!]
\centering
\footnotesize
\begin{minipage}{0.22\linewidth}
\centerline{\includegraphics[width=\textwidth]{ 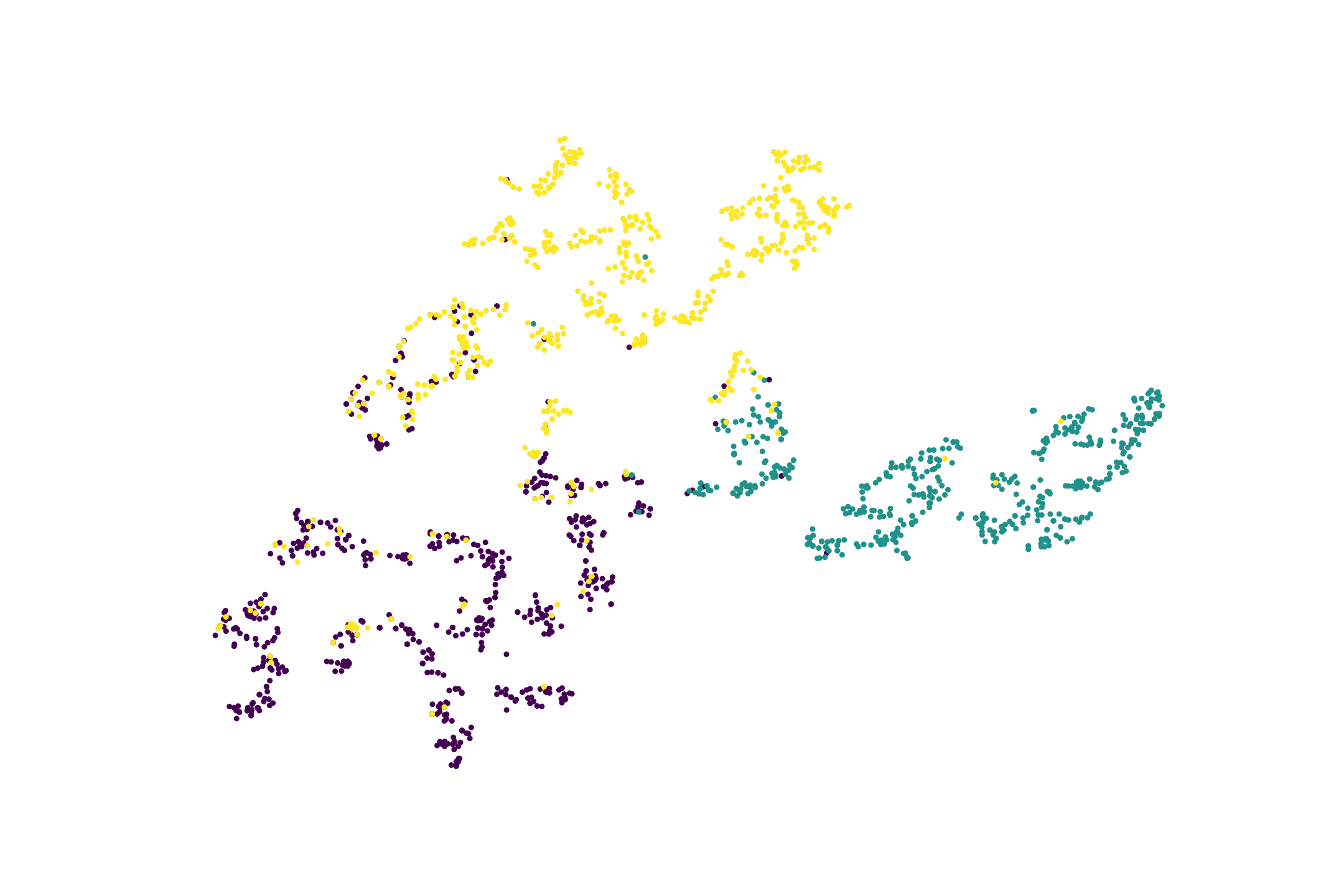}}
\vspace{3pt}
\centerline{\includegraphics[width=\textwidth]{ 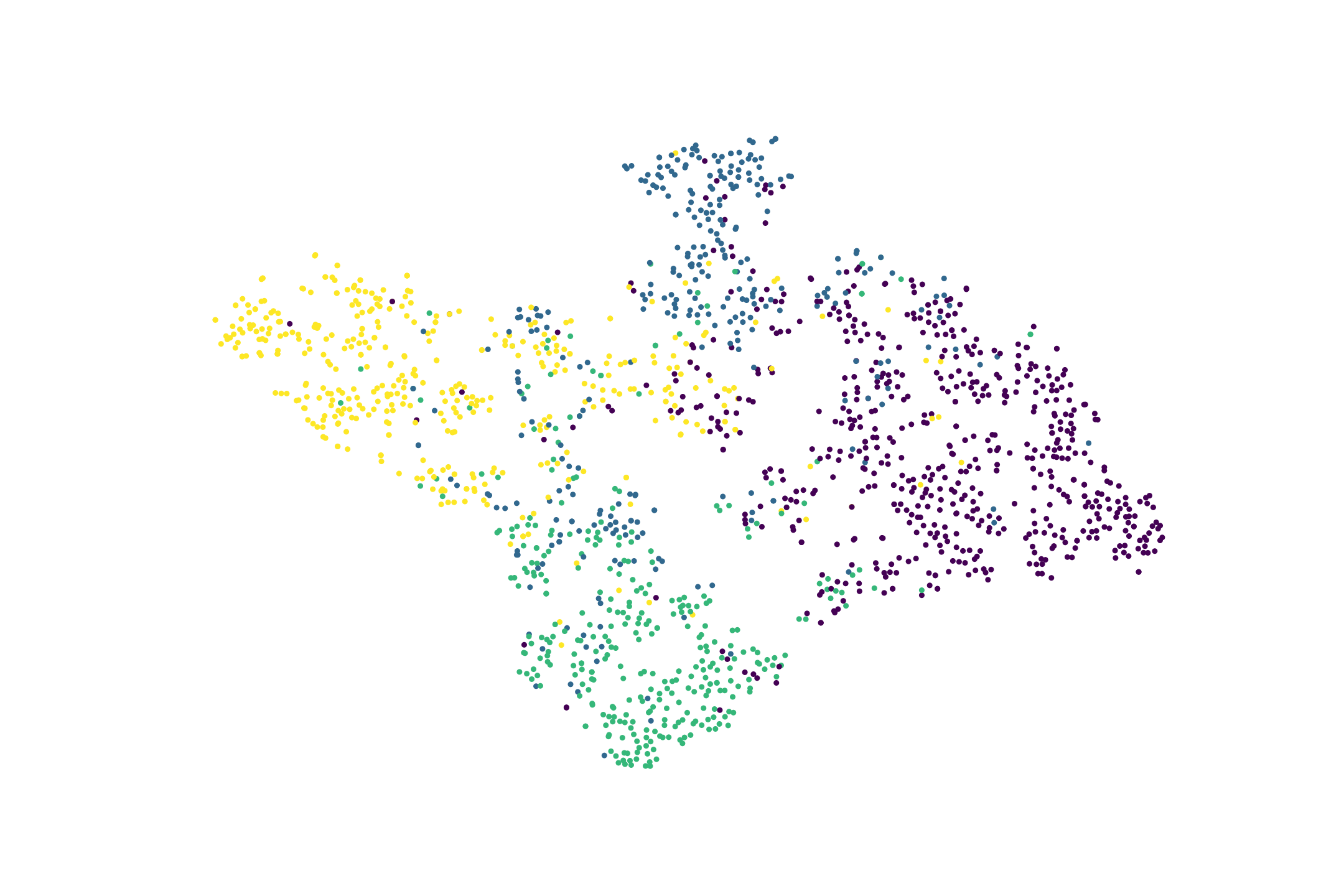}}
\vspace{3pt}
\centerline{GCN}
\end{minipage}
\begin{minipage}{0.22\linewidth}
\centerline{\includegraphics[width=\textwidth]{ 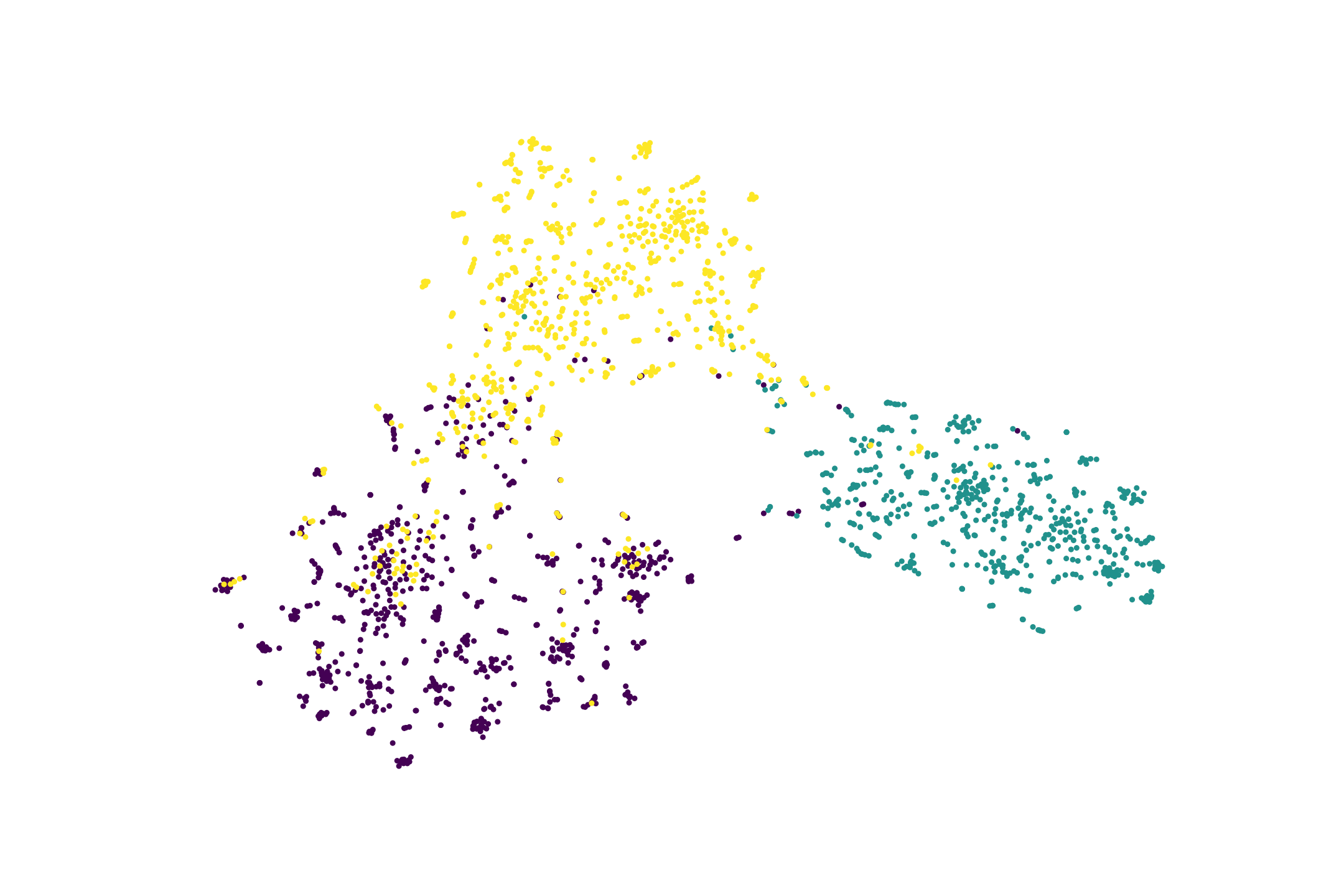}}
\vspace{3pt}
\centerline{\includegraphics[width=\textwidth]{ 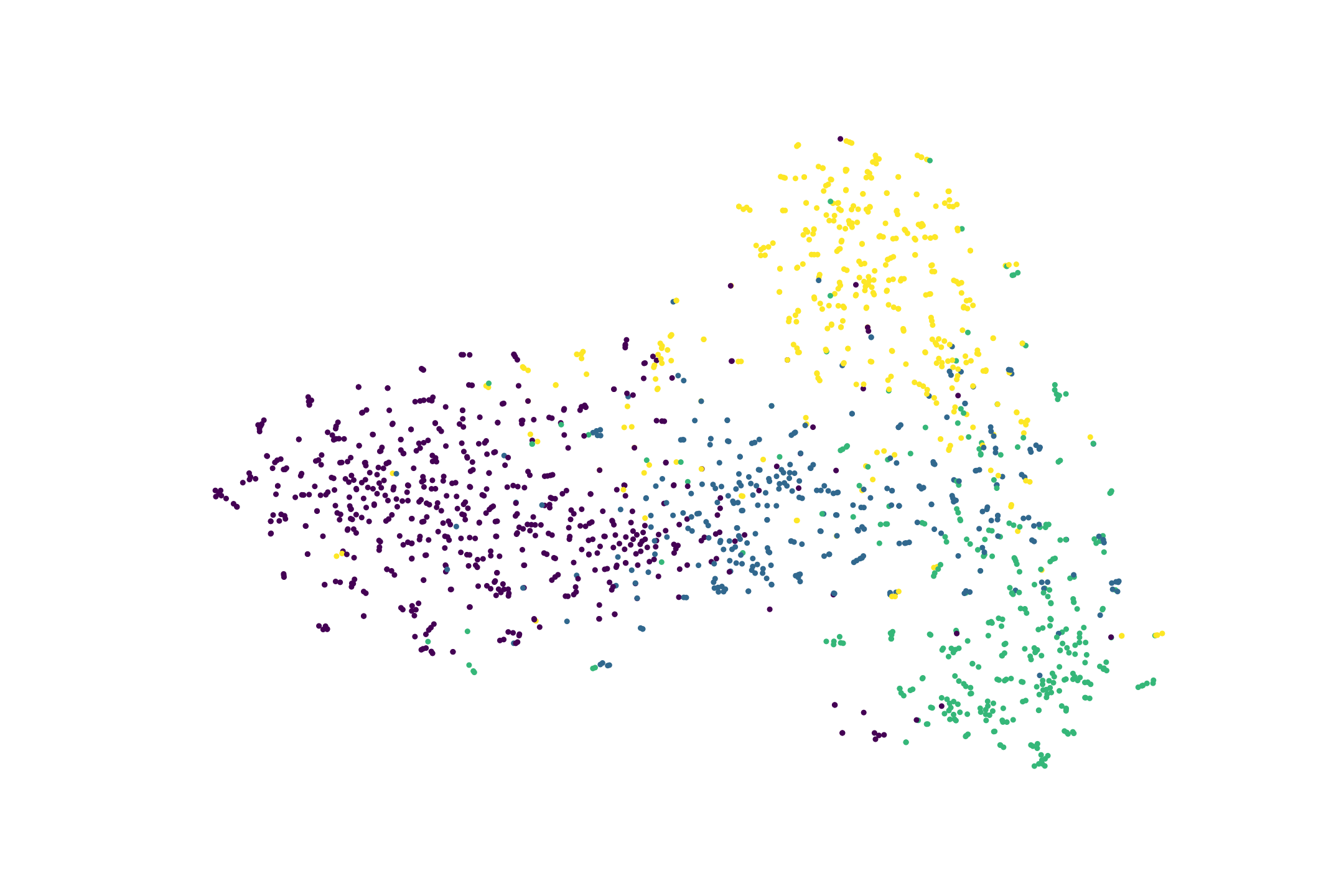}}
\vspace{3pt}
\centerline{FAGCN}
\end{minipage}
\begin{minipage}{0.22\linewidth}
\centerline{\includegraphics[width=\textwidth]{ 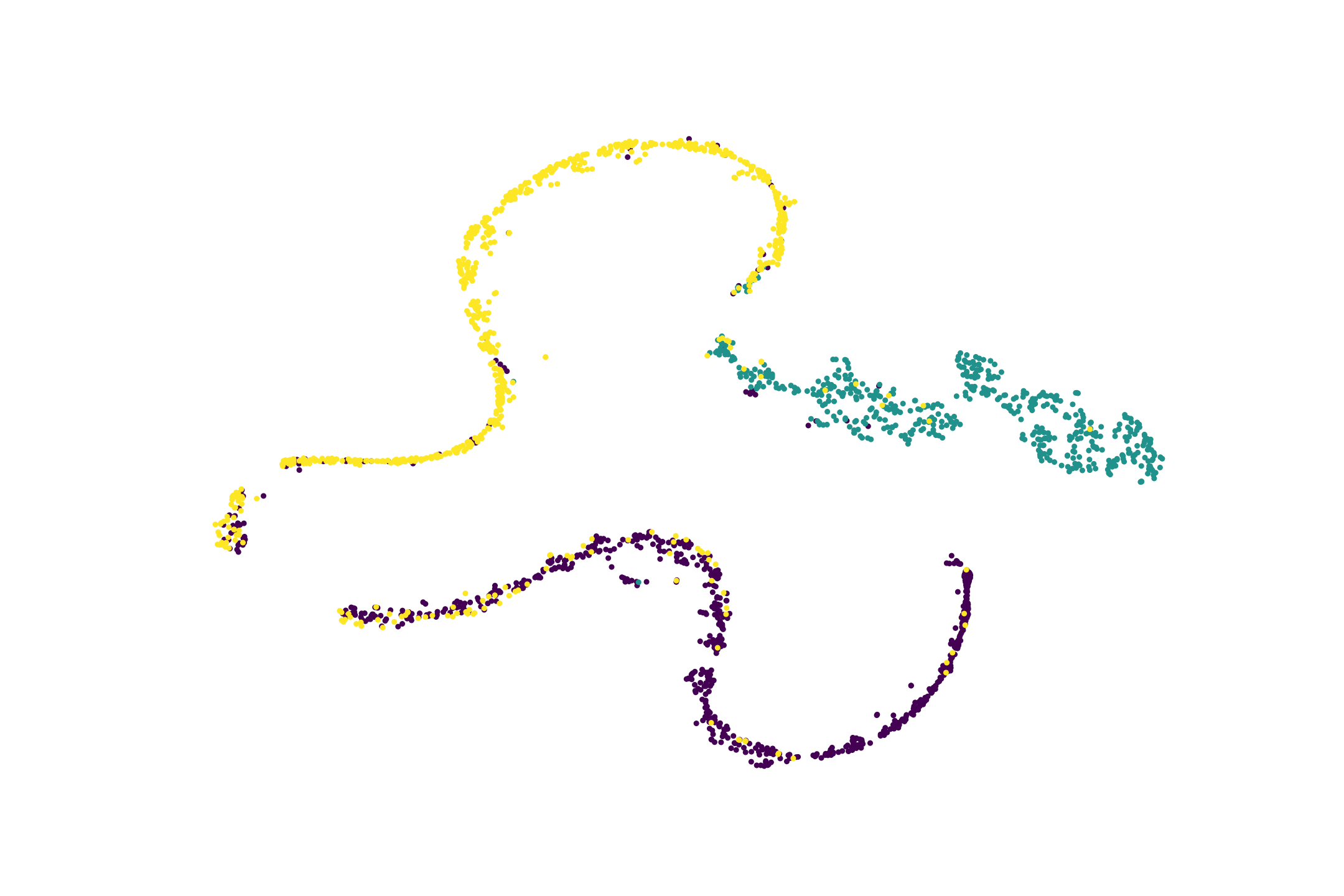}}
\vspace{3pt}
\centerline{\includegraphics[width=\textwidth]{ 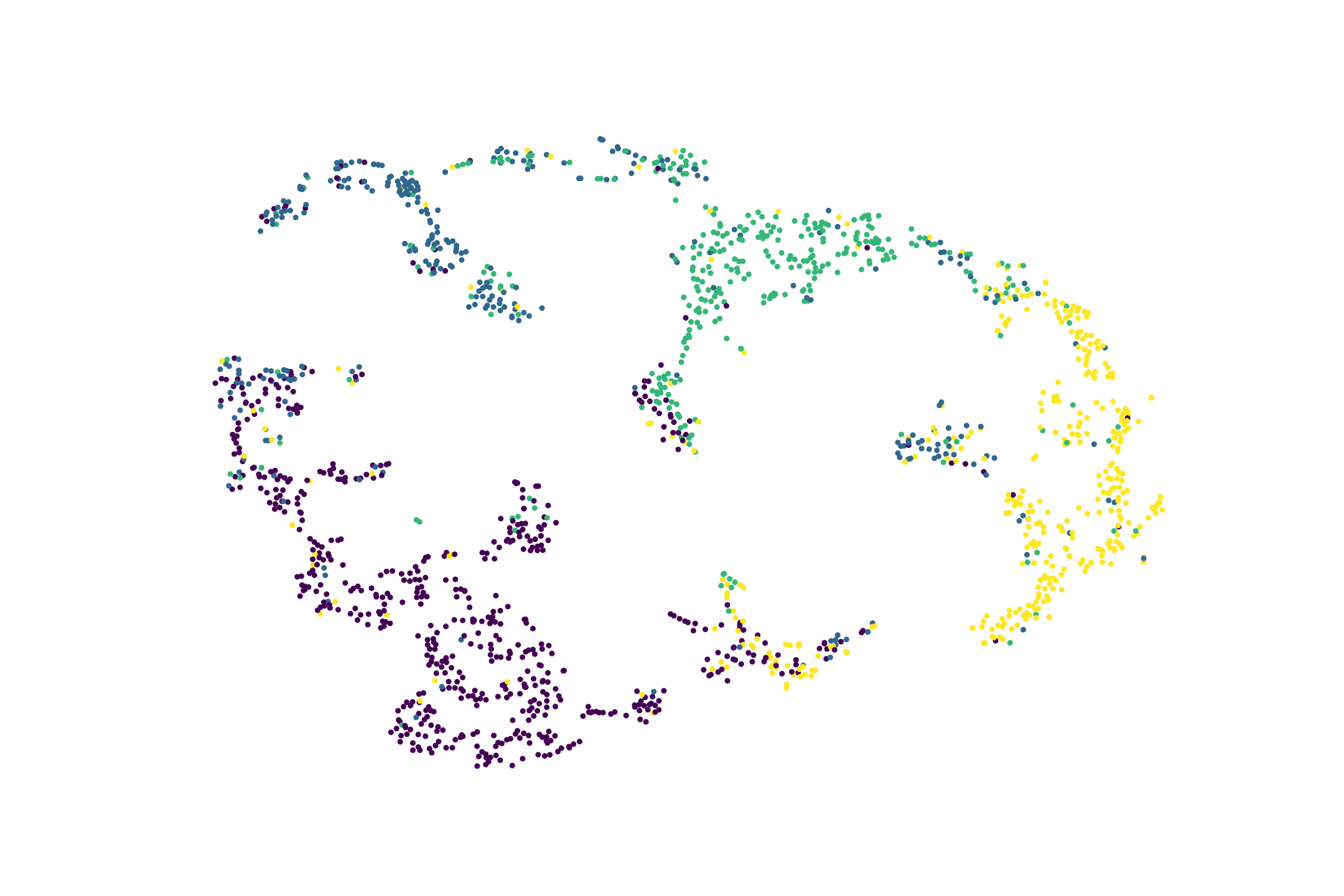}}
\vspace{3pt}
\centerline{MixupForGraph}
\end{minipage}
\begin{minipage}{0.22\linewidth}
\centerline{\includegraphics[width=\textwidth]{ 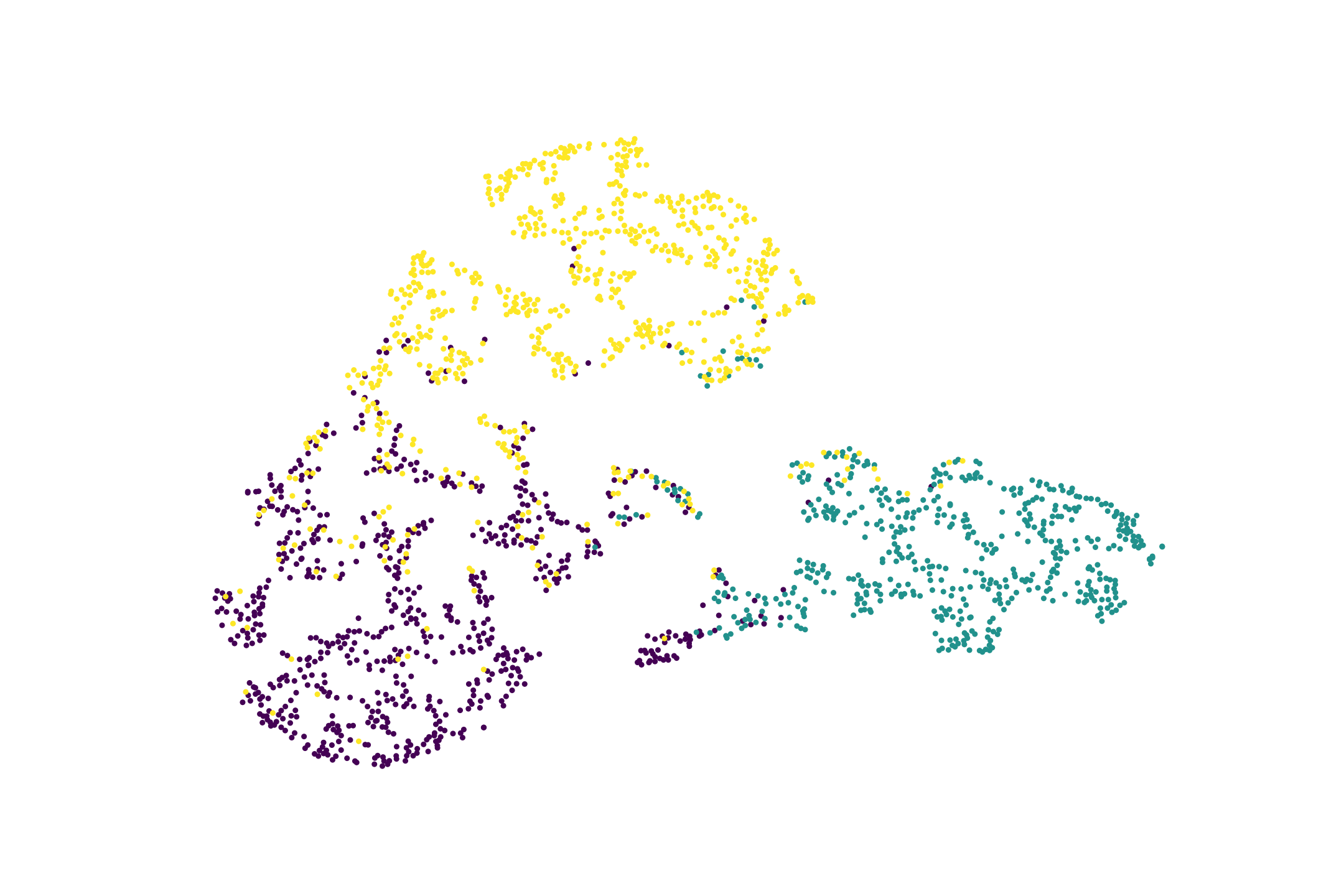}}
\vspace{3pt}
\centerline{\includegraphics[width=\textwidth]{ 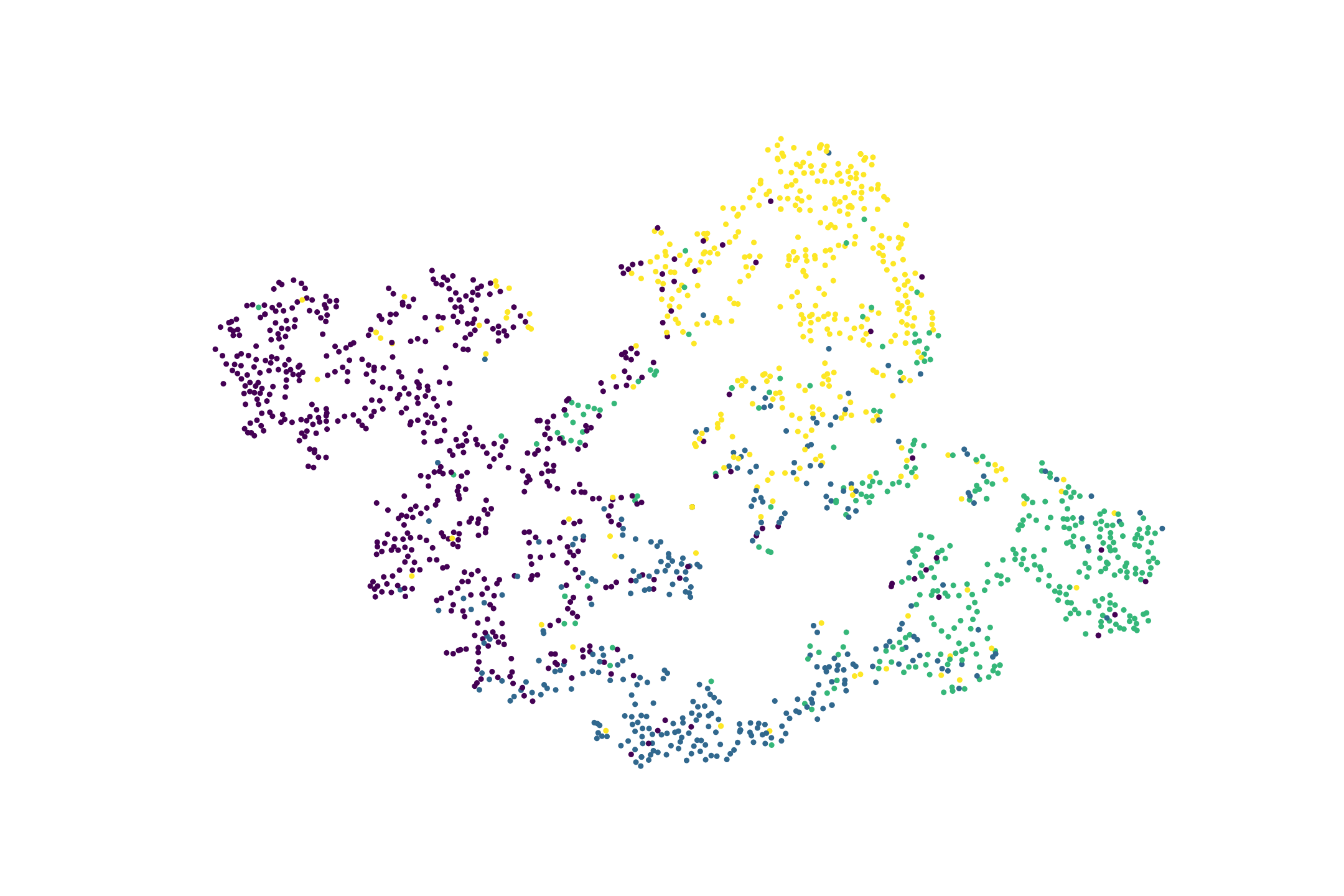}}
\vspace{3pt}
\centerline{PPNP}
\end{minipage}
\\
\begin{minipage}{0.22\linewidth}
\centerline{\includegraphics[width=\textwidth]{ 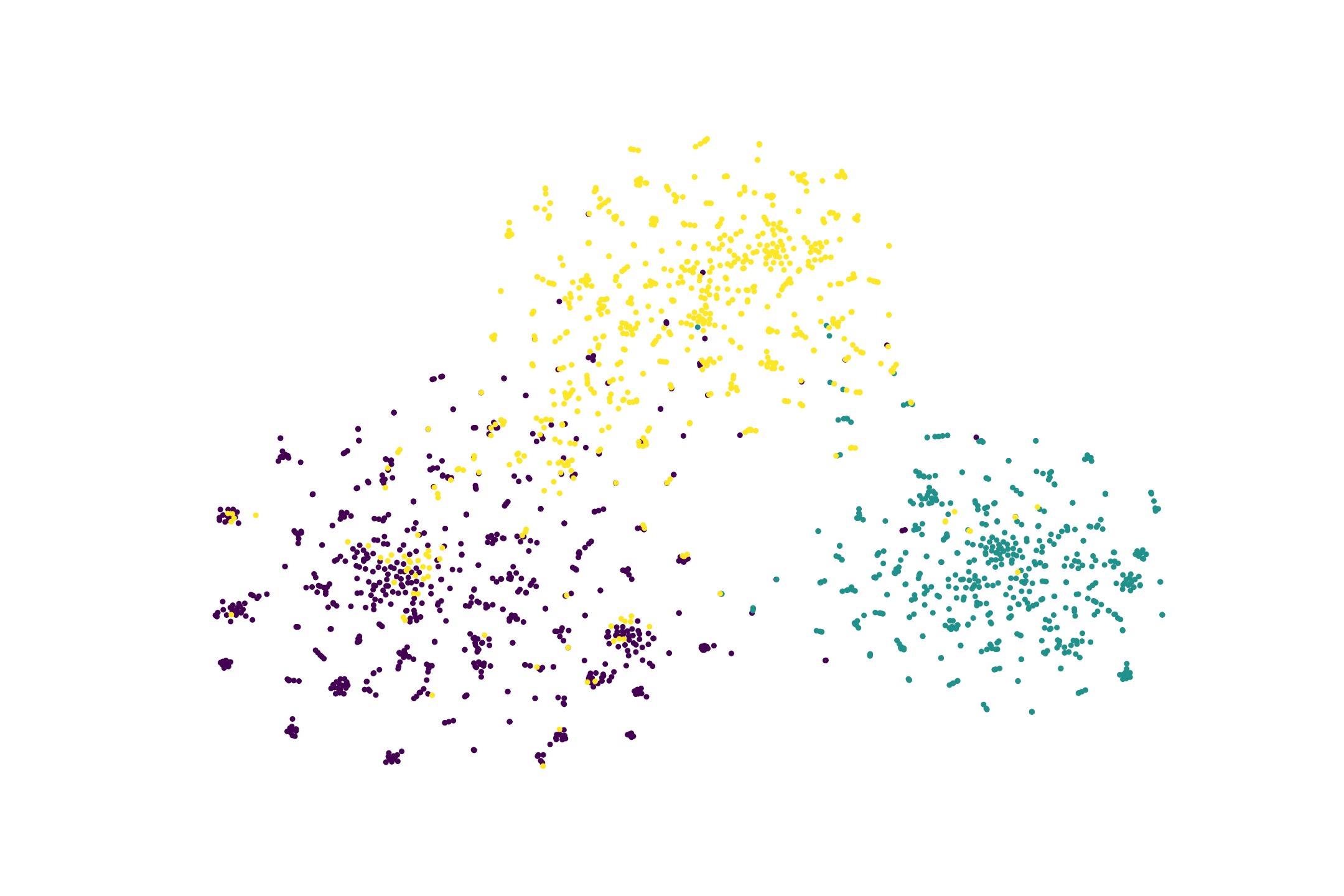}}
\vspace{3pt}
\centerline{\includegraphics[width=\textwidth]{ 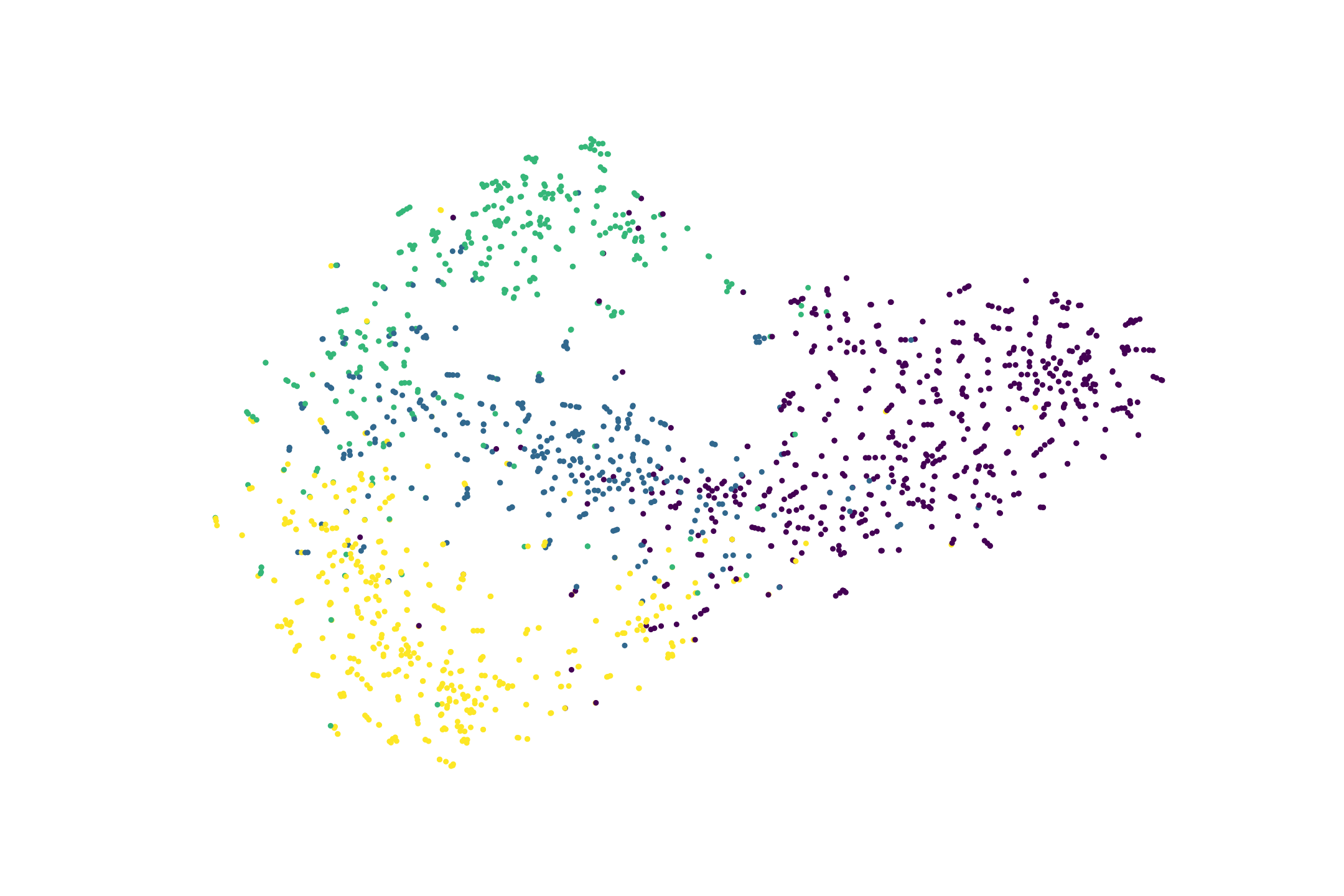}}
\vspace{3pt}
\centerline{GWNN}
\end{minipage}
\begin{minipage}{0.22\linewidth}
\centerline{\includegraphics[width=\textwidth]{ 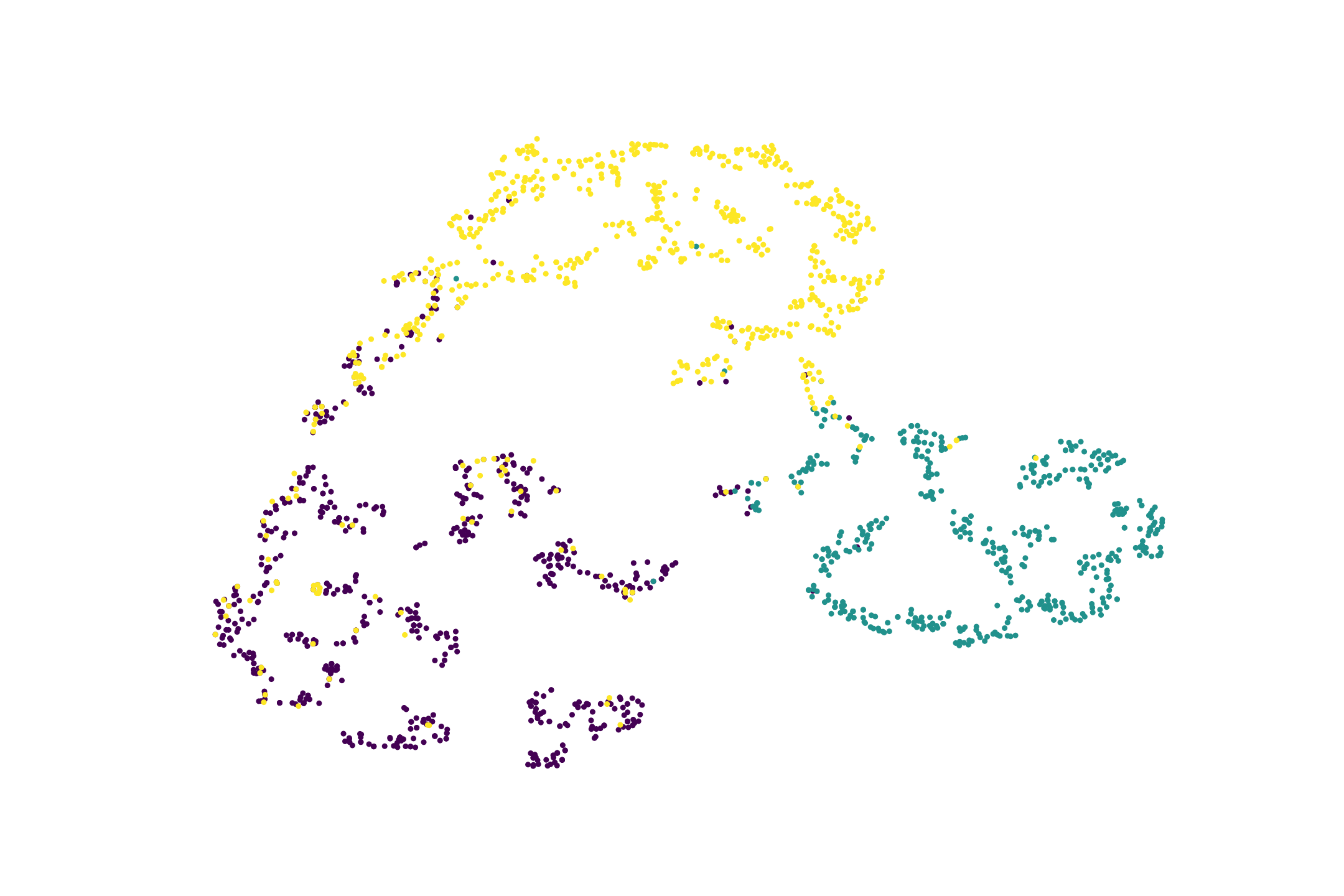}}
\vspace{3pt}
\centerline{\includegraphics[width=\textwidth]{ 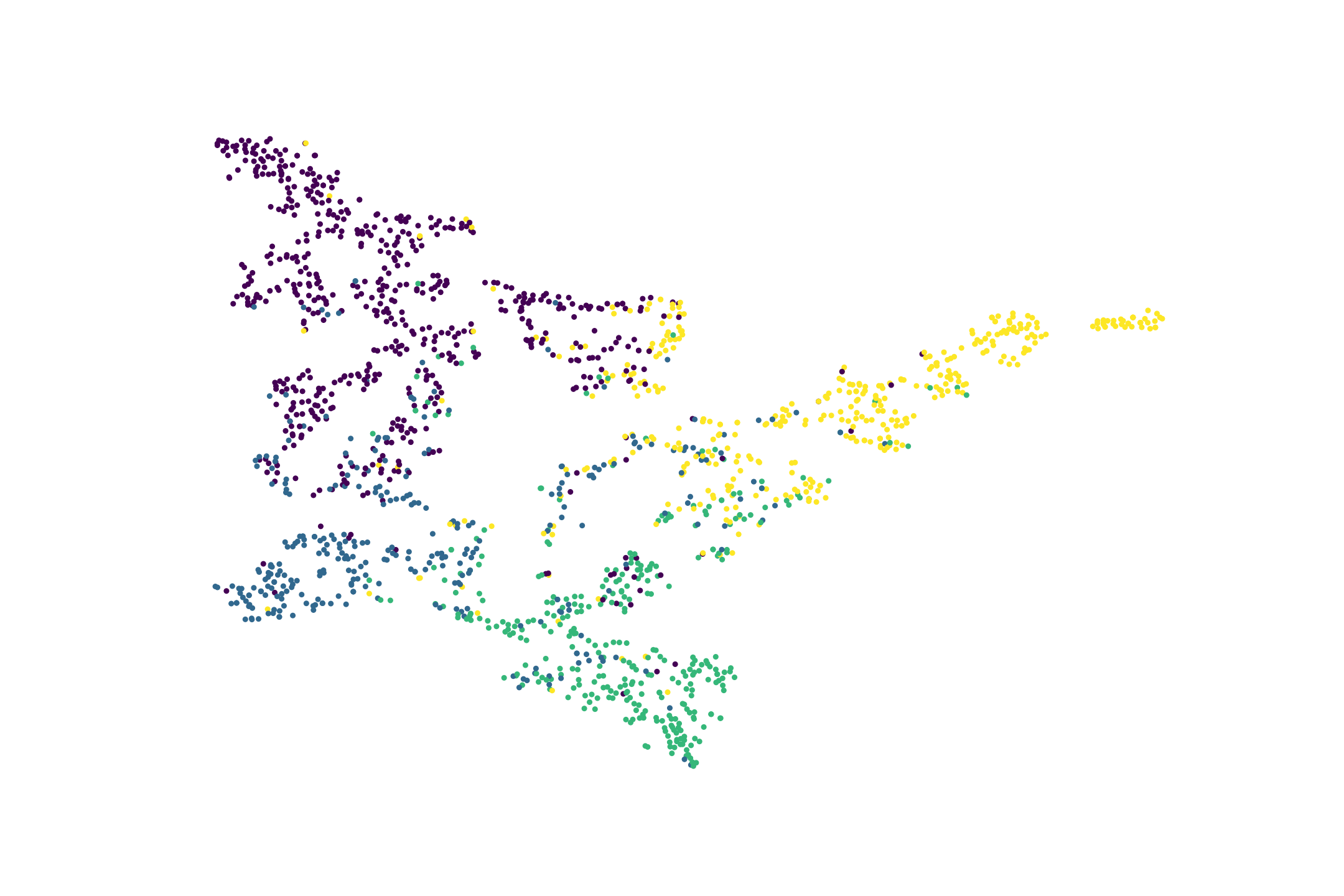}}
\vspace{3pt}
\centerline{ADAGCN}
\end{minipage}
\begin{minipage}{0.22\linewidth}
\centerline{\includegraphics[width=\textwidth]{ 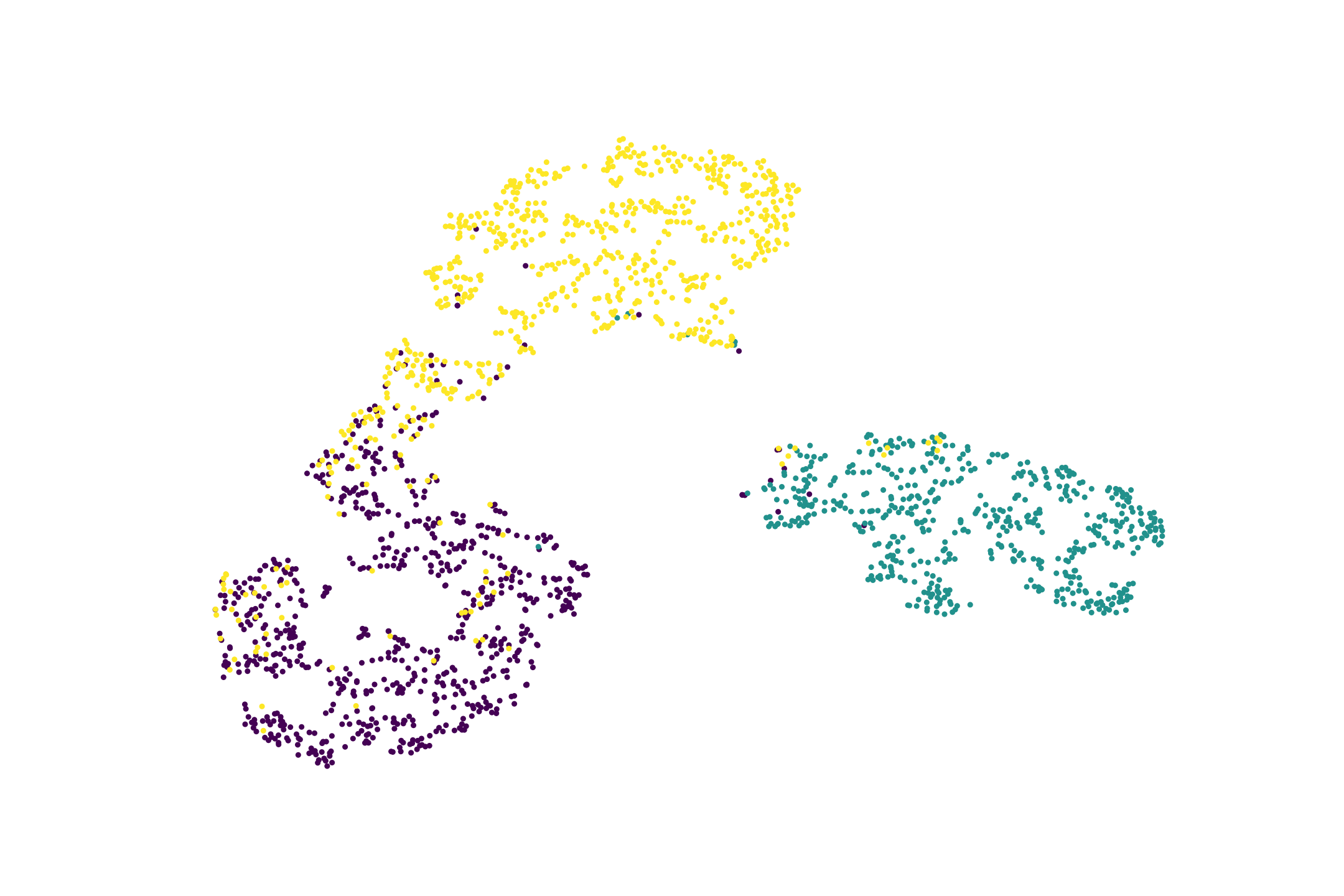}}
\vspace{3pt}
\centerline{\includegraphics[width=\textwidth]{ 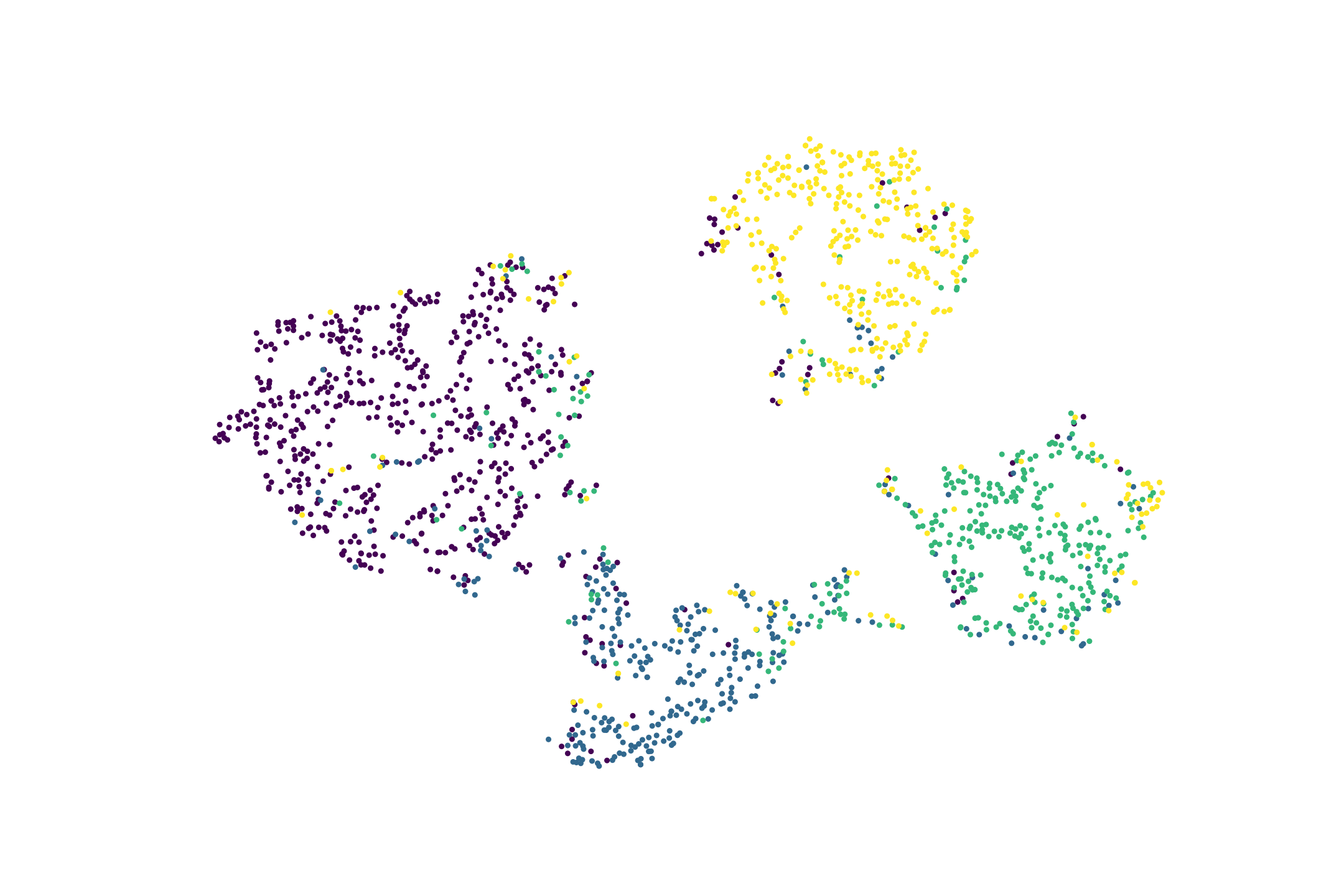}}
\vspace{3pt}
\centerline{GANN(Ours)}
\end{minipage}
\caption{$t$-SNE visualization of seven algorithms on two benchmark datasets. The first row corresponds to the ACM dataset, and the second row relates to the DBLP dataset.}
\label{tsne}  
\end{figure}

\begin{figure}[h!]
\centering
\footnotesize
\begin{minipage}{0.22\linewidth}
\centerline{\includegraphics[width=\textwidth]{ 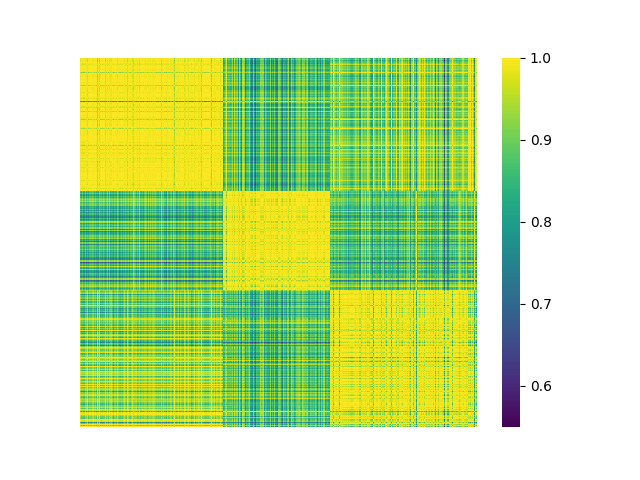}}
\vspace{1.5pt}
\centerline{\includegraphics[width=\textwidth]{ 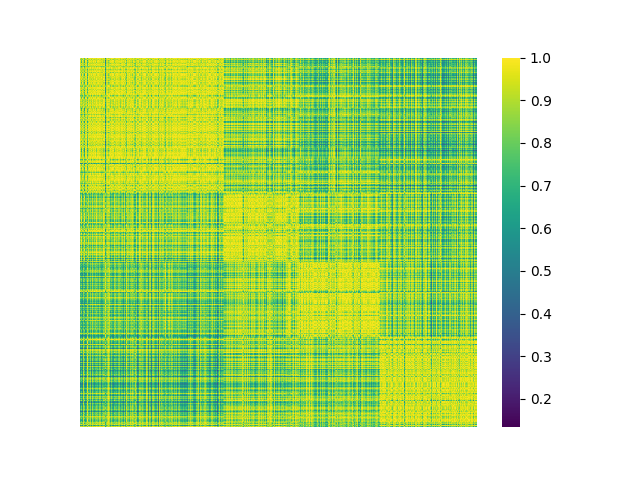}}
\vspace{1.5pt}
\centerline{GCN}
\end{minipage}
\begin{minipage}{0.22\linewidth}
\centerline{\includegraphics[width=\textwidth]{ 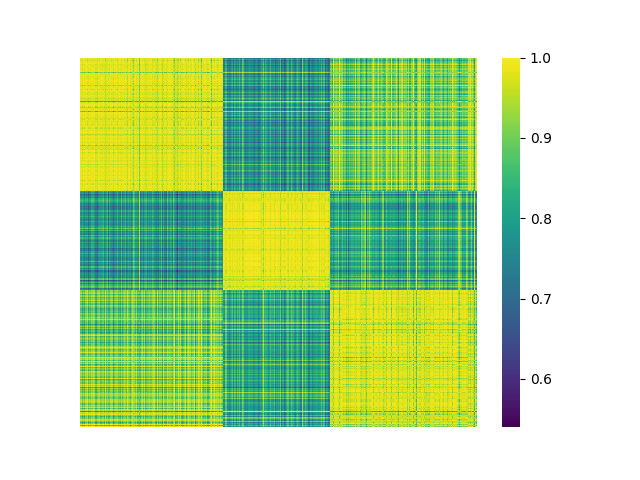}}
\vspace{1.5pt}
\centerline{\includegraphics[width=\textwidth]{ 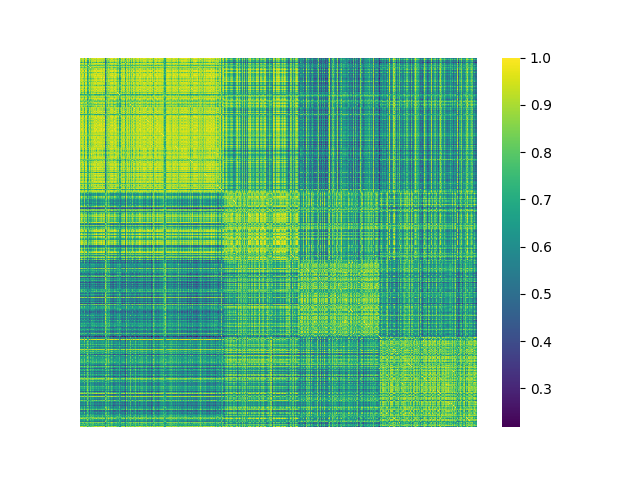}}
\vspace{1.5pt}
\centerline{FAGCN}
\end{minipage}
\begin{minipage}{0.22\linewidth}
\centerline{\includegraphics[width=\textwidth]{ 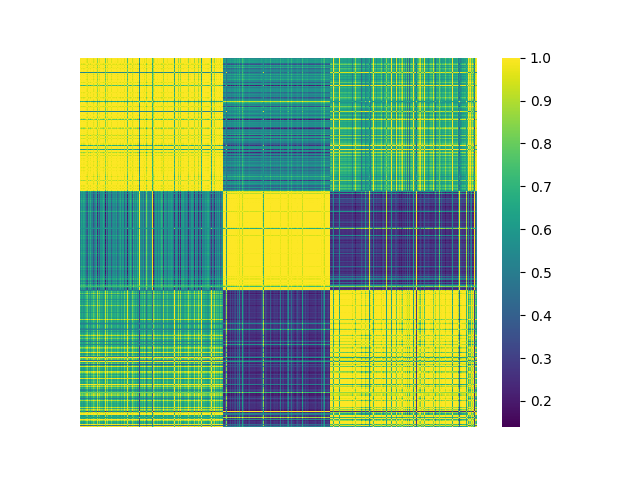}}
\vspace{1.5pt}
\centerline{\includegraphics[width=\textwidth]{ 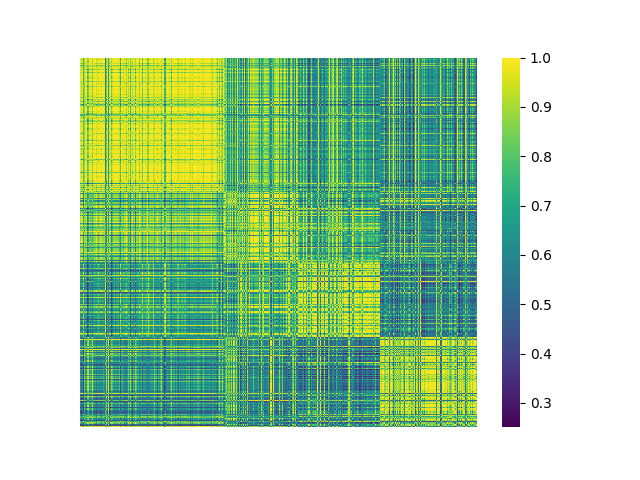}}
\vspace{1.5pt}
\centerline{MixupForGraph}
\end{minipage}
\begin{minipage}{0.22\linewidth}
\centerline{\includegraphics[width=\textwidth]{ 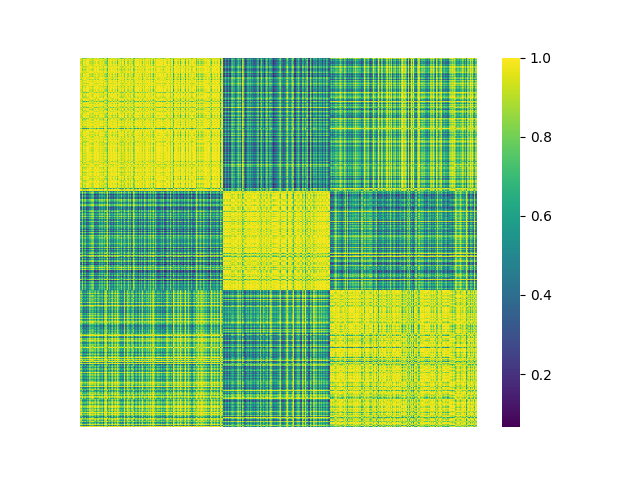}}
\vspace{1.5pt}
\centerline{\includegraphics[width=\textwidth]{ 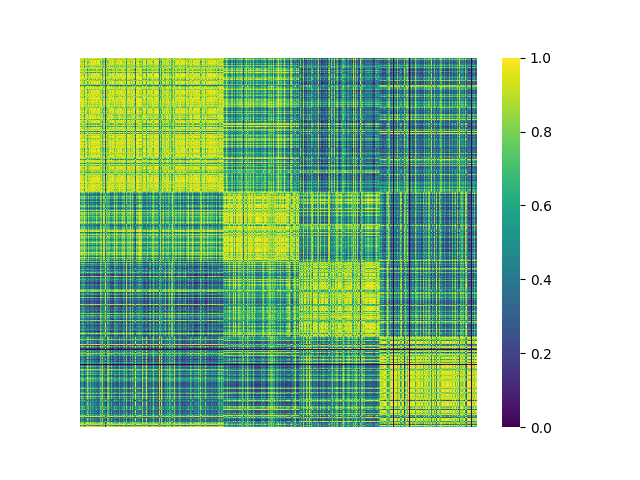}}
\vspace{1.5pt}
\centerline{PPNP}
\end{minipage}
\\
\begin{minipage}{0.22\linewidth}
\centerline{\includegraphics[width=\textwidth]{ 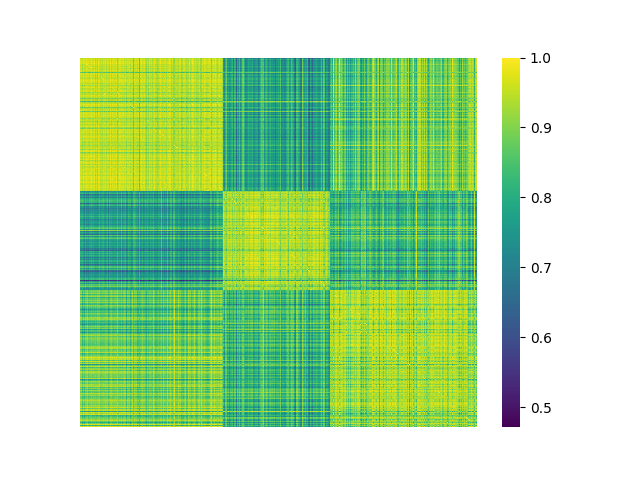}}
\vspace{1.5pt}
\centerline{\includegraphics[width=\textwidth]{ 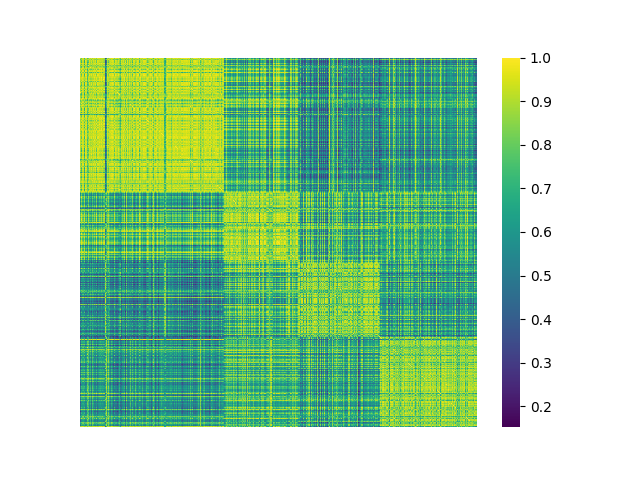}}
\vspace{1.5pt}
\centerline{GWNN}
\end{minipage}
\begin{minipage}{0.22\linewidth}
\centerline{\includegraphics[width=\textwidth]{ 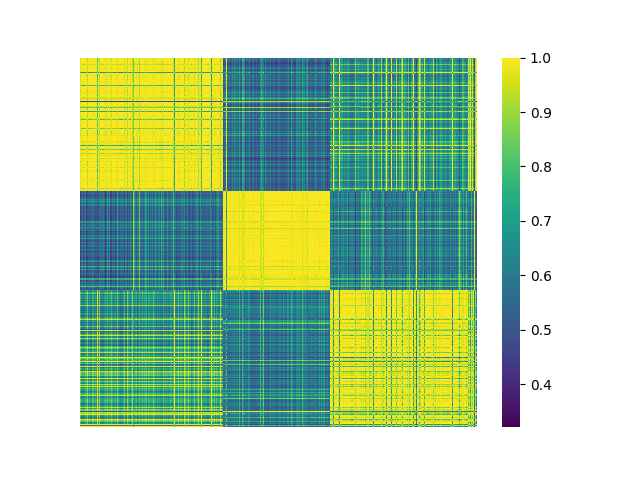}}
\vspace{1.5pt}
\centerline{\includegraphics[width=\textwidth]{ 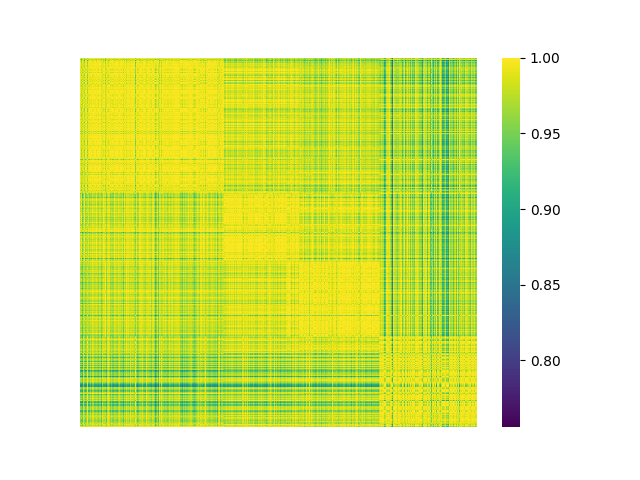}}
\vspace{1.5pt}
\centerline{ADAGCN}
\end{minipage}
\begin{minipage}{0.22\linewidth}
\centerline{\includegraphics[width=\textwidth]{ 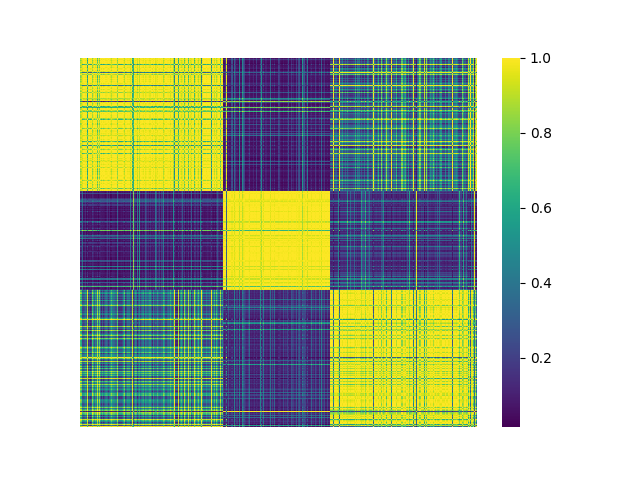}}
\vspace{1.5pt}
\centerline{\includegraphics[width=\textwidth]{ 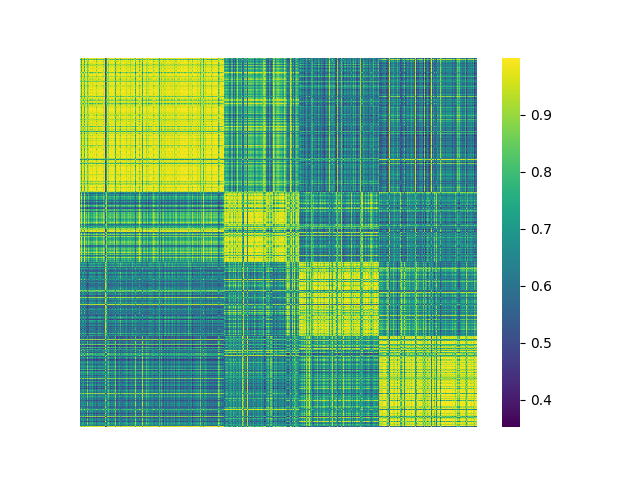}}
\vspace{1.5pt}
\centerline{GANN(Ours)}
\end{minipage}
\caption{Visualization of sample similarity matrices on two datasets. The first row and second row correspond to ACM and DBLP, respectively.}
\label{sim}
\end{figure}

\subsection{Visualization Experiments}
In this section, we compare the models visualized from two different angles to further demonstrate GANN's superiority. We select seven models, including GCN, FAGCN, MixupForGraph, PPNP, GWNN, ADAGCN, and GANN, on ACM and DBLP datasets.
1) To show the cluster division of node representations, we display the visualization output by $t$-SNE algorithm\cite{tsne} in Figure \ref{tsne}. From the results, it can be seen that the plots generated by GCN, PPNP and GWNN, the boundaries between categories are not obvious and the clusters are blended together, indicating that the quality of node representation is not high. It's obvious that our proposed GANN can improve the quality of node representations by effectively distancing the categories from each other, and then better classify node. 2) To visualize the oversmoothing issue in the graph node classification task, we generate heat maps of the sample similarity matrices corresponding to node representations. All samples are sorted by category so that examples from the same cluster are near to one another. Figure \ref{sim} demonstrates that the oversmoothing issue is particularly severe in GCN and ADAGCN. Our proposed GANN is capable of learning more unique node representations, which significantly alleviates the problem of oversmoothing.

\section{5 \hspace{0.3cm} Conclusion}
In this paper, we study the semi-supervised problem on graphs and offer GANN, a simple and efficient framework. In GANN, we propose three alignment rules: feature alignment rule, cluster center alignment rule and minimum entropy alignment rule. We perform ablation experiments and are able to demonstrate well that GANN can fully exploit the feature and higher-order neighbor information of the data to generate more discriminative node representations, and alleviate the problem of oversmoothing. On eight datasets, we demonstrate that GANN outperforms thirteen other relevant models and superior in time and space to most models. In particular, when using a very small labeling rate, our model guarantees the best results when using only $7$ labeled samples per cluster.
In conclusion, the simple and successful principles given in GANN may serve as a paradigm for future GNN models, particularly those using semi-supervised learning. By augmenting the data, we intend to further enhance the scalability of GANN in future work. Also we will consider adding decoders in the second half of the model to try the effect of other GNN models combined with MLP. Equally essential work is the processing of large-scale datasets, and we will include this part of the data in the code details and check how well the model works. 
\clearpage
\thispagestyle{empty}

\bibliography{new}
\clearpage
\thispagestyle{empty}

\end{document}